\documentclass[journal]{IEEEtran}
\usepackage{blindtext}
\usepackage{graphicx}

\ifCLASSINFOpdf
\else

\fi

\usepackage[cmex10]{amsmath}
\usepackage{algorithmic}
\usepackage[tight,footnotesize]{subfigure}
\let\MYorigsubfigure\subfigure
\renewcommand{\subfigure}[2][\relax]{\MYorigsubfigure[]{#2}}
\usepackage{graphicx}
\DeclareMathOperator*{\argmax}{arg\,max}
\DeclareMathOperator*{\argmin}{arg\,min}
\newcommand{\norm}[1]{\left\lVert#1\right\rVert}
\newlength\myindent
\newcommand\bindent{%
  \begingroup
  \setlength{\itemindent}{\myindent}
  \addtolength{\algorithmicindent}{\myindent}
}
\newcommand\eindent{\endgroup}

\begin{document}
%
\title{HEX and Neurodynamic Programming}
%
%
%

\author{Debangshu Banerjee,  
\thanks{D. Banerjee is with the Stochastic Systems Lab
, Indian Institute of Science, Bangalore,
KA, 560012 }
}

%



\maketitle

\begin{abstract}
Hex is a complex game with a high branching factor. For the first time Hex is being attempted to be solved without the use of game tree structures and associated methods of pruning. We also are abstaining from any heuristic information about Virtual Connections or Semi Virtual Connections which were previously used in all previous known computer versions of the game. The H-search algorithm which was the basis of finding such connections and had been used with success in previous Hex playing agents has been forgone. Instead what we use is reinforcement learning through self play and approximations through neural networks to by pass the problem of high branching factor and maintaining large tables for state-action evaluations. Our code is based primarily on NeuroHex. The inspiration is drawn from the recent success of AlphaGo Zero.  
\end{abstract}


%
\IEEEpeerreviewmaketitle

\section{Introduction to the game of HEX}
HEX is a two-person board game invented by Danish mathematician Piet Hien in 1942 at Neils Bohr Institute and independently by Nobel-Laureate John Nash in 1948 at Princeton University \cite{wikiHex}. Hex is a strategy game, that gives high significance to the players' decision-tree style thinking and situational awareness in determining the outcome \cite{wikiHex}. The board consists of hexagonal grids called cells arranged to form a rhombus. Typically $n \times n$ boards are used with championships being held on $11 \times 11$ boards. Players alternate placing markers on unoccupied cells in an attempt to link the opposite sides of their board in a continuous chain.
\subsection{Hex and Mathematics}
The current area of research related to Hex can be found in areas of topology, graph and matroid theory, combinatorics, game theory, computer heuristics, artificial intelligence \cite{wikiHex}. Hex is a connection game \cite{wikiHex} in which players try to connect the opposite edges using their pieces by employing strategical decisions. It is a Maker-Breaker type positional game \cite{wikiHex} which is described as : the Maker wins by capturing all positions that lead to a connection, while if the game ends with all positions claimed but the Maker not having won, it implies the Breaker has won. This brings to the first theorem about Hex that Hex cannot end in a draw \cite{wikiHex}, \cite{HEX}. Though John Nash is credited for the proof \cite{HEX} there is no supposting material. The proof is credited to David Gale, American Mathematician and Economist, University of California, Berkley \cite{HEX}.\par
Hex is a perfect information game \cite{wikiHex}, which means that each player while making a decision is perfectly informed about all events that have previously happened. The condition of perfect information and no-draw makes Hex a determined game which means that for all instances of the game there is a winning strategy for one of the players \cite{wikiHex}.\par
This brings us to the most famous theorem about Hex, that the first player has the winning strategy \cite{HEX}. This theorem was proved by John Nash using a strategy-stealing argument \cite{HEX}. \par
 David Gale's proof that Hex cannot end in a draw uses a graph theory lemma \cite{HEX}. A simple graph with nodes having degree at most 2 can be shown to be a union of disjoint sets of either isolated nodes, simple cycles or simple paths. By constructing a subgraph from the original Hex graph by including only special edges, the nodes of the subgraph is shown to have degree at most two. The lemma is then applied to show existence of simple paths between opposite edges \cite{HEX}.
 That Hex cannot end in a draw has been shown equivalent to the existence of the two dimensional Brouwer fixed-point by David Gale \cite{HEX}. The Brouwer fixed-point theorem was key to proving the existence of Nash Equilibrium \cite{HEX}.

\subsection{HEX and Computers}
With Nash's theorem about first person having the winning strategy, people were interested in solving the game of Hex by hand as well as by computers \cite{arneson2010solving} .\par
The first hex playing machine was constructed in 1950 by Claude Shannon and E. F. Moore \cite{ShannonHex}. It was an analog machine. A two-dimensional potential field was set up corresponding to the playing board. With black trying to connect the top and bottom edges, black pieces and the top and bottom edges were given negative charge and white pieces were given positive charge along with the two sides. The move to be made was specified by a certain saddle point in the potential field. The machine performed reasonably well and won about 70 percent of the games with opening moves.\par
However, until recently, machines had never been able to challenge human capability \cite{arneson2010solving}. To understand the complexity of Hex,note that the game-tree search techniques that are applicable in chess becomes less useful in Hex because of the large branching factor. A standard $11\times 11$ board in Hex has on an average 100 legal moves compared to 38 for chess \cite{ANSHELEVICH2002101}. \par
Another blow to the hopes of finding good algorithms to solve Hex came in 1976, when Shimon Even and Robert Tarjan showed the problem of determining which player has a winning strategy in Shannon's Switching Game is PSPACE complete \cite{even1976combinatorial} and finally in 1981 Stefan Reisch proved PSCPACE completeness for $N \times N$ Hex \cite{reisch1981hex}. These results indicate that there is little chance of having a polynomial time algorithm which could find a winning strategy.\par
Still there are many results about the solution of Hex which deserve a mention : \par
Smaller games upto $4 \times 4$ have been completely solved by hand. Piet Hein, one of the inventors, commented that interesting positions start to arise from $5 \times 5$ boards. Martin Gardner, a popular mathematics writer briefly discussed winning opening moves on board sizes of $5 \times 5$. In 1995, Bert Enderton solved all $6 \times 6$ openings which was later verified in 2000 by Van Rijswijck \cite{arneson2010solving} the founder of Queenbee, a HEX program that won silver at London 2000 Computer Games Olympiad \cite{van2000bees}.\par
In 2001 Yang solved the $7 \times 7$ board by hand for a number of openings \cite{yang2001decomposition}, \cite{Anewsolutionfor77HexGame}. He used a decomposition technique where each board configuration of Hex can be viewed as a sum of local games. Each local game can be solved using a particular strategy. By decomposing any Hex board into a sum of local games and using particular sets of strategies on each local game a win is ensured. Yang used 41 such local game patterns in his 12 page case analysis.\par
However the major breakthrough in computer Hex came in 2000 by Anshelevich.
V.V. Anshelevich, a Russian American physicist, gave the first Hex playing algorithm \cite{ANSHELEVICH2002101}, \cite{anshelevich2000game}. The computer program developed was called HEXY, winner of the gold medal at the 2000 Computer Olympiad, London. Anshelevich's contribution was phenomenal in that he built the foundation on which all later strong Hex playing algorithms would be built.\par
Anshelevich et al used the concept of "sub games" following the decomposition idea introduced by Yang. They proposed an algorithm called H-Search. In their program of HEXY an alpha beta search is used on a game-tree with each node, representing a board configuration, has its value estimated using an evaluator function based upon the H-Search algorithm. \par
Though the H-Search algorithm is theoretically insufficient, the method has been used in all major Hex playing agents with success \cite{van2002search}, \cite{hayward2004solving}, \cite{hayward2005solving}, \cite{henderson2009solving}. In 2003 Hayward et al , University of Alberta, Computer Hex Research Group, solved all 49 $1^{st}$ place openings on $7 \times 7$ boards. Since then the success of Hex playing agents has steadily increased with machines like SIX, WOLVE, MONGOOSE, SOLVER etc, all developed by the Computer Hex Research Group at the University of Alberta \cite{uAlberta}. Each of these has won medals in the Computer Game Olympiads along with solving for all $1^{st}$ place opening positions for boards up to sizes $9 \times 9$.
The strongest Hex playing engine till date is known as MOHEX \cite{arneson2010monte}, \cite{cazenave2010monte}, \cite{huang2013mohex}. MOHEX was written by Ryan Hayward and Phillip Henderson, Computer Hex Research Group, University of Alberta in 2007-10. It uses Monte Carlo Tree Searches with Upper Confidence Bounds to balance exploration versus exploitation and prunes nodes via the Virtual Connection Method discussed along with other improvements and heuristics. MOHEX has recently won the gold medal on $11 \times 11$ and $13 \times 13 $ boards in the 2015 International Computer Games Association, Computer Games Olympiad held at Netherlands. Till date centre opening winning strategy for $10 \times 10$ board has been shown \cite{uAlberta}. Currently the Computer Hex Research Group at the University of Alberta is associated with maintaining all these engines and carrying forward their research in the game of Hex.\par
In a completely different setting, David Silver et al, at Google DeepMind in 2016 presented ALPHAGO \cite{silver2016mastering} which beat a human world champion, Ke Jie, three times out of three in the ancient Chinese board game of GO. Their success was contributed to the use of a large data set of GrandMaster moves which were used for supervised training of a neural network followed by self-play and reinforcement learning for improvement. Inspired by this, Kenny Young, Ryan Hayward, Computer Hex Research Group, University of Alberta developed Neurohex \cite{young2016neurohex} , a reinforcement learning agent trained for Hex. Since our work is highly based on this paper we will explain this in detail.\par
In 2017, \cite{silver2017mastering} the same team of ALPHAGO, developed ALPHAGO ZERO, a Hex playing engine which is architecturally simpler than ALPHAGO. Also it is completely devoid of any supervision and relies only upon reinforcement learning. ALPHAGO ZERO has been reported to beat ALPHAGO 100 to 0, calling into question human strategies employed in HEX used to train ALPHAGO.

\section{Our Contribution}
Our contribution is mostly experimental, focusing on various Reinforcement Learning algorithms. We focus mainly on algorithms that aim to decrease the Projected Bellman Error. We also compare these algorithms with those which tend to minimize the traditionally used Value Error. We focus on both Critic Methods as well as Actor-Critic-methods to solve control problems of Reinforcement Learning.  We compare results both with and without the use of supervised pre-training of network weights using a database of optimal moves. We also use a Recurrent Neural Network based architecture and compare the same with a Convolutional Neural Network based architecture used in Neurohex \cite{young2016neurohex}. Our final results indicate that using a Recurrent Neural Network might not be better suited in comparison to the existing Convolution Neural Networks when used in the contexts of positional games like Hex \cite{young2016neurohex}. We also provide a new optimization technique that can be used in conjunction with function approximation and Reinforcement Learning. We hope that our optimization technique aims to reduce the Projected Bellman Error, though theoretical analysis remains to be done. Empirical results indicate that our method of optimaization gives better results than traditional algorithms like Q-learning and SARSA, which aim to decrease the Value Error. \par
Our paper is organized as follows : Section III introduces ideas from Reinforcement Learning, Section IV deal with Experiments and Section V deals with Results. Finally Section VI and VII discuss Conclusion and Future Work respectively.

\section{Reinforcement Learning and Neuro-Dynamic Programming \cite{sutton1998reinforcement}}
Reinforcement Learning as the name suggests, focuses on learning through reinforcement of positive actions. Actions are termed positive if they are beneficial. Actions have consequences and beneficial actions are those which have better consequences. At a particular time instance the learning agent encounters a state of the environment. The environment is the particular problem setting which we aim to solve. In our case the environment is the Hex problem. Having encountered an environmental-state the agent takes an action. Actions are dependent on the state encountered. Not all actions are legal in every state. The action changes the environmental state from one to another. The transitions can be deterministic or stochastic. The transition also results in either a reward or a loss signal depending on the environment. For example leaning slightly on a bicycle will cause it to turn but leaning too much  will result in a fall. Normally we can provide a very simple algorithm which does not involve any learning : choose actions which give the highest rewards. But what happens usually is that we are more interested in looking for long term goals. For example if I paddle fast on a bicycle I will not fall so I can choose to paddle fast, but I will also get tired quickly so I may not reach my destination! Here my goal is reaching the destination without falling and that would mean I optimize my pedalling.\par
Hence, we have an objective: to maximize the cumulative reward one receives, starting from an initial state and upon taking a sequence of actions. Most reinforcement learning algorithms try to achieve this, i.e. assess how good a state (in our case a board configuration) is in the long term by calculating the expected cumulative reward received by taking a sequence of actions. For example we try to maximize the chance of winning while having encountered a particular board-configuration by trying different action sequences. This can be easily done by simulation and maintaining a list of all environmental states. We choose a state and run a simulation based on a model of the environment following a particular predefined set of actions. The model produces the state transitions and reward signals. We add up the rewards to get a return. This return is a sample of the value function of that state following that particular set of actions. (If state evolution is random we would need to take the average of the returns based on a certain number of simulations). We might also have different action sequences. We can calculate the value function by following each and every action sequence to get the value of the state corresponding to different action sequences. Finally we can choose the one action sequence which gives the highest return for that state. We can carry this out for every state of the environment to identify which action sequence gives the highest return when in a particular state.\par   
Instead of running an entire simulation we can also use backed up values of the states themselves to assess the values of other states.$$ v_\pi(s) = \mathbf{E}_\pi\big [\sum_{t=0}^TR_{t+1} | s_0 = s \big ]$$ 
$$ = \mathbf{E}_\pi\big [ \sum_{t=0}^{T-1} R_{t+1} + v_\pi(s_T) | s_0 = s \big ]$$
$$ = \mathbf{E}_{s'\sim\pi(s'|s,\pi(s))}\big [ R_1 + v_\pi(s_1)|s_0 = s \big ]$$where $\pi$ is the particular action sequence $a_0, a_1,...a_{T-1}$. One can initialize $v_\pi$ to some value for all states and apply the above recursive formula iteratively to converge to a solution of the above equation (Much like a Gauss-Seidel approach to solving a linear system of equations). \par
With this we understand the dynamic portion of the Neurodynamic Programming name. To understand the first part we have to look from the neural network function approximation perspective. In problems where there are many states it is reasonably difficult to assess the value function of all of these states mainly because of storage constraints and general inefficiency. So we try to build a parameterized function with the number of parameters much less than the number of states. Our hope is that with a state as input to the function, we will get $v_\pi(s)$. Obviously, because of the large difference in the number of inputs and the number of parameters we will not get a perfect $v_\pi(s)$. But, we are not looking for perfection. What we are looking for is an approximation, because what we really want to find is an optimal action sequence or policy. This will be found as that action at a given state which gives the maximum state-action value out of all actions applicable in that particular state. Thus relative correct comparisons of state-action values for different actions in a state would do our job. So we can apply function approximation with some confidence. We optimize the network parameters so as to decrease the error $\norm{\hat{v}(s,w) - v_\pi(s)}$. This is similar to all supervised learning techniques \cite{sutton1998reinforcement}. The only difference that arises in reinforcement learning is that the target $v_\pi$ is also unknown. So we need to approximate $v_\pi(s)$ either as a sample $\sum_{t=0}^TR_{t+1}$ obtained from simulation of model following a policy $\pi$ or using bootstrapped values $R_{t+1} + v_\pi(s_{t+1})$.\par
With this we understand the meaning of the term Neurodynamic Programming and also get the basic idea of policy evaluation, which is estimating $v_\pi(s)$ given a policy or an action sequence $\pi$. Policy Evaluation is one step of the Policy Iteration Algorithm. The 2nd step is of course the Policy Improvement algorithm. This is what we are primarily interested in : To find a policy better than what we have now \cite{sutton1998reinforcement}.\par
To summarize :
\begin{itemize}
    
    \item Estimate $v_\pi(s)$ for each state $s$ using a policy $\pi$ 
    \item For each $s$ recompute $\pi(s)$ as the action $a$ which maximizes $v_{\pi(s)}(s)$ 
    \item Continue until $\pi(s)$ does not change for any state $s$.

\end{itemize}

This is the general Reinforcement Learning Technique known as Policy Iteration. One question that arises immediately is what policy should we begin with? We can start with any random policy \cite{sutton1998reinforcement}. In our work we always start with a random policy.
\begin{enumerate}
    \item Assign $\pi_0(s)$ as : For state $s$ choose any action $a$ with uniform probability over all legal actions in state $s$.
    \item Repeat for all states $s$.
\end{enumerate} 

However, when we don't have a model of the environment to explicitly calculate the expectation $\mathbf{E}_{s'\sim\pi(s'|s,\pi(s))}\big [ R_1 + v_\pi(s_1)|s_0 = s \big ]$ and we have to rely only on samples of simulated trajectory we run into the problem of exploration vs exploitation \cite{Bertsekas:1996:NP:560669}, which more often than not results in bad estimates of the optimal policy. To see clearly why this is so let us maintain returns not only for the state but also on a particular action taken in that state. This does not greatly change our notation. Instead of estimating $v_\pi(s)$, where in state $s$ we took an action $\pi(s)$, we estimate $v_a(s)$, the return followed by taking a particular action $a$ in state $s$ and following policy $\pi$ for all other states. So, $v_a(s_t) = R_{t+1} + v_\pi(s_{t+1})$. Note that if $\pi(s) = a$, this would have exactly corresponded to $v_\pi(s)$. Usually $\mathbf{Pr}\{\pi(s) = a\} > 0 , a \in \mathcal{A}(s)$. This slight deviation is not of great significance in our problem because for games such as Hex, checkers or GO the separateness of state and state-action pair is not that extreme and we can combine both in an after-state \cite{sutton1998reinforcement}. Maintaining returns for a state-action pair has its advantages when using samples to estimate optimal policy\cite{sutton1998reinforcement}.\par
Imagine that we take the initial policy to be random and run the Policy Iteration Algorithm. We also assume that we do not calculate the true expected returns of each state but instead rely on  sample trajectories. We also assume the number of states and actions corresponding to each state are small and can be stored within a table itself.
\begin{enumerate}
    \item Initialize $v_a$ to zero as a table of the size corresponding to the total number of states and actions $a$ legal in each state $s$, and $\pi_0$ is random.
    \item Run a simulated trajectory $s^t_0, s^t_1,....s^t_T$ based on policy $\pi_t$ which is as we have defined.
    \item \quad For each state $s$ encountered in the trajectory update $v_a(s_t) = R_{t+1} + v_b(s_{t+1})$ where $a$  and $b$ are the actions chosen according to $\pi_t(s_t)$ and $\pi_t(s_{t+1})$. 
    \item \quad For each $s$ recompute $\pi_t(s)$ as the action $a$ which maximizes $v_a(s)$ 
    \item Go to step 2. Continue till $\pi_t(s) = \pi_{t-1}(s)$ for all states $s$.
\end{enumerate}
With the above procedure, actions which are not picked at the first policy, that is transitions of the nature $s_t \xrightarrow{\text{a}} s_{t+1}$ which are not encountered, will not result in a computation of $v_a(s)$ where $a$ is a valid action that could have been taken in state $s$. Therefore, even if $a$ was the optimal action to be taken in state $s$, it would not be included during the policy improvement step of $\pi_{t+1}(s) = \argmax_{a' \in \mathcal{A}(s) / a} v_{a'}(s)$. \par
To circumvent this problem we note that had we allowed in step 2 the policy $\pi_0$ to visit all $S^A$ actions for all states, we could have repeated step 3 for a large number of iterations before reaching step 4, we would not have had this problem. For then, $v_a(s)$ would have been estimated quite nicely for each state-action pair. However in a true online algorithm we want to do step 3 and step 4 after each $s_t \xrightarrow{a} s_{t+1}$ transition. To do this one trick is to use $\epsilon$-greedy policy \cite{sutton1998reinforcement}. An $\epsilon$-greedy policy is a stochastic policy all the way. Which is to say it never converges to a deterministic policy, but the policy does improve and reaches an $\epsilon$-optimality. Here while choosing an action based on the policy we don't always choose the greedy action with respect to the state-action value. Instead we do it randomly, that is with probability $\frac{\epsilon}{|\mathcal{A}(s)|}$ we choose a non-greedy, random action while w.p $ 1 - \epsilon + \frac{\epsilon}{|\mathcal{A}(s)|}$ we choose the greedy action w.r.t the current state-action value \cite{sutton1998reinforcement}. We do this every time we simulate a trajectory and update according to the state-action-reward-state-action tuple encountered. This way one can be assured that the optimal policy is achieved at the end of the procedure \cite{sutton1998reinforcement}.\par
Another method to by pass this constraint is to use Policy Gradient techniques \cite{sutton1998reinforcement}. Unlike $\epsilon$-greedy policies, where there is no explicit policy without the value function, policy gradient techniques focus on an explicit description of the policy itself. Policy Gradient techniques use a parameterized function approximation of a policy and improve the policy by optimizing the parameters so as to maximize the return from a state following that policy.Thus, given an environmental state as input to the function the output is a probability distribution over all actions allowable in that state and parameters are updated in a direction that increases $v_{\pi_\theta}(s)$, which is $\nabla v_{\pi_\theta}(s)$. Parameters of the policy-network are updated based on gradient-ascent techniques. More often than not actor-critic methods are used, where not only value functions of states are updated, but also the policy is improving based on newer estimates of the value functions. Policy Gradient techniques can converge to a local optimum unlike $\epsilon$-greedy policies where there is always an $\epsilon$-probability of choosing a random action.\par
The final topic of interest is the error used in conjunction with function approximation \cite{AnAnalysisofTemporalDifferenceLearning}. As noted earlier, $\norm{\hat{v}(s,w) - v_\pi(s)}$ is the error one tries to reduce. But there are a number of facets to this problem. First, the error that we are trying to reduce is $\mathbf{E}_\pi\big[ v_\pi(s) - \hat{v}(s,w) \big ]^2$, where it is assumed that each state is distributed in the steady-state according to the on-policy distribution $\pi$ which is being used to interact with the environment. If $v_\pi(s)$ was known we could have used samples instead of the expectation and applied gradient descent techniques to optimize the parameters. On the other hand, if $v_\pi(s)$ is unknown, we can use estimates from simulated trajectories. One such estimate is $\sum_{t=0}^TR_{t+1}$, that is the entire reward obtained through the length of the trajectory starting from state s. Using this estimate for $v_\pi(s)$ also leads to good use of stochastic gradient techniques to optimize parameters. But what happens when we use bootstrapped estimates, like $R_{t+1} + v_\pi(s_{t+1})$? Here, one can get $v_\pi(s_{t+1})$ by the approximation $\hat{v}(s_{t+1},w)$. The update would be proportional to the error, $R_{t+1} + \hat{v}(s_{t+1},w) - \hat{v}(s_{t},w)$.Note that, for a true gradient descent, the direction should be along the negative of $\nabla_w\hat{v}(s_{t+1},w) - \nabla_w\hat{v}(s_{t},w)$. However, what one usually does is what is known as semi-gradient descent, which is to say ignore the effects of the changing weight vector on the target \cite{sutton1998reinforcement}, i.e., use the direction as $-\nabla_w\hat{v}(s_{t},w)$. Semi-Gradient techniques have also been used successfully with on-policy bootstrapped targets as estimates for learning through function approximation.\par
The assumption that the states are distributed according to the transition probabilities of the Markov Chain does not necessarily hold true as in most cases of control through $\epsilon$-greedy policies. These form the basis of off-policy control in Markov-Decision-Processes. Here, states are distributed according to a behaviour policy, while the transitions about which we are interested are those due to a target policy. The return from a state as simulated due to a behaviour policy is therefore not a true estimate of the return one would have obtained if the target policy would have been followed. One can get around this problem by applying importance sampling ratio to the returns under the behaviour policy. Thus $v_\pi(s)$ can be approximated as $\rho\sum_{t=0}^TR_{t+1}$ with $\rho$ as the importance-sampling ratio, and then apply stochastic gradient as earlier. The problem arises when one tries to use bootstrapped values along with semi-gradient techniques. There are counter examples \cite{baird1995residual},  \cite{AnAnalysisofTemporalDifferenceLearning} and a work by Richard Sutton and Mahmood \cite{DBLPjournalscorrSuttonMW15} which show that such techniques would diverge when used in conjunction with function approximation. To address this problem, the target had to be examined with its relation to function approximation.\par
It can be assumed that the class of functions defined by a given architecture cannot exactly represent $v_\pi(s)$ for any $s \in S$ \cite{sutton1998reinforcement}. Thus the best we can do is find parameters so that the error, $\big[ v_\pi(s) - \hat{v}(s,w) \big ]^2$, is a minimum on average. When $v_\pi(s)$ is estimated as the cumulative reward over the entire trajectory, then the solution obtained is the best solution one could arrive at for approximating $v_\pi(s)$. However, when we use bootstrapped values the solution is often a bad approximation. To understand this observe that when we apply the Bellman Operator, $T^\pi$ \cite{Bertsekas1996NP560669} on a value function $\hat{v}$ representable by the function architecture, the result, $\mathbf{E}_\pi\big[ R_{t+1} + \hat{v}(s_{t+1},w)\big ]$, is a new value function, generally not representable by the architecture. On repeated application of the Bellman-Operator, the value-functions converge to the true $v_\pi$. However, with function approximation, each intermediate value function is projected back into the function space, before applying the Bellman Operator. This leads to different estimates for the true value function $v_\pi$. Since, $v_\pi$ is the only fixed point of the Bellman Operator \cite{Bertsekas1996NP560669}, that is to say, it is the only point where the Bellman Error, $\mathbf{E}_\pi\big[ R_{t+1} + v_\pi(s_{t+1}) - v_\pi(s_{t}))\big ]$ is zero, one could very easily see that the Bellman Error can never be reduced to 0 while using function approximation unless the true value function is representable via the parameterized class of value functions (obtained from using function approximation) itself. However, if we were to project the Bellman Error in the function space one could find a value function where the Projected-Bellman Error is 0 \cite{sutton1998reinforcement}. These areas were studied by Richard Sutton, Hamid Maei, Doina Precup, Shalabh Bhatnagar, David Silver, Csaba Szepesvari, Eric Wiewiora and they have given algorithms which are based on actual gradient techniques which tend to minimize the Projected Bellman Error \cite{NIPS20083626}, \cite{Sutton2009FGM15533741553501}, \cite{bhatnagar2009convergent}, \cite{maei2010gq}, \cite{maei2011gradient}.\par
There is an additional point we would like to make, that of eligibility traces. We will not go into the full details here, except to note that there exists a parameter $\lambda$ which can be used to tune how much of the returns are to be accumulated before using bootstrapped values of states encountered in sample paths. $\lambda = 0$ corresponds to just a single transition, whereas $\lambda = 1$ implies the entire cumulative reward one obtains at the end of the trajectory. One can choose any $\lambda$ between 0 and 1 \cite{sutton1998reinforcement}.

\section{Experiments}
We run all experiments on a $3 \times 3$ Hex board. The reason for doing this is we know what the optimal game play should be in $3 \times 3$ and moreover we believe that once the key elements for ensuring optimal play have been identified, scaling to larger systems would be comparatively simpler than without having done so. Large scales might impose other problems like that of exploration, but we defer the issue to future work. \par
Each board configuration has been modelled as a $6 \times 3 \times 3$ Boolean 3d array \cite{young2016neurohex}. The 6 channels correspond to the following\par
0 : a 3 $\times$ 3 array with positions for red stones marked as true.\par 
1 : a 3 $\times$ 3 array with positions for blue stones marked as true.\par
2 : a 3 $\times$ 3 array with positions for which red stones form a continuous unbroken chain attached to the east wall marked as true.\par
3 : a 3 $\times$ 3 array with positions for which red stones form a continuous unbroken chain attached to the west wall marked as true.\par
4 : a 3 $\times$ 3 array with positions for which blue stones form a continuous unbroken chain attached to the north wall marked as true.\par
5 : a 3 $\times$ 3 array with positions for which blue stones form a continuous unbroken chain attached to the south wall marked as true.\par
Along with this for each of the 6 channels, each side of the $3 \times 3$ array is augmented by 2 units to form the red and blue edges, thus resulting in a $7 \times 7$ array. The $0^{th}$ channel is thus augmented with 2 columns at the beginning and end marked as true and 2 rows at the top and bottom marked as false, the inner $3 \times 3$ board remains as it is. The $1^{st}$ channel is augmented with 2 columns at the beginning and end marked as false and 2 rows at the top and bottom marked as true, the inner $3 \times 3$ board remains as it is. The $2^{nd}$ channel is augmented with 2 columns at the beginning marked as true and 2 columns at the end marked as false and 2 rows at the top and bottom marked as false, the inner $3 \times 3$ board remains as it is. The $3^{rd}$ channel is augmented with 2 columns at the beginning marked as false and 2 columns at the end marked as true and 2 rows at the top and bottom marked as false, the inner $3 \times 3$ board remains as it is. The $4^{th}$ channel is augmented with 2 columns at the beginning and at the end marked as false and 2 rows at the top marked as true and 2 rows at the bottom marked as false, the inner $3 \times 3$ board remains as it is. The $5^{th}$ channel is augmented with 2 columns at the beginning and at the end marked as false, 2 rows at the top marked as false and 2 rows at the bottom marked as true, the inner $3 \times 3$ board remains as it is. Thus the input is pre-processed into a $6 \times 7 \times 7$ three-dimensional array \cite{young2016neurohex}. \par
The neural network architecture used has 4 layers \cite{young2016neurohex}. The $1^{st}$ layer is a 2d convolution layer acting on the pre-processed input, the $2^{nd}$ layer is another 2d convolution layer acting on the output of the $1^{st}$ layer. The $3^{rd}$ layer is also a 2d convolution layer which acts on the $2^{nd}$ layer output and the $4^{th}$ layer is a sigmoid layer acting on the product of a matrix and the output of the $3^{rd}$ layer to produce a $9 \times 1$ vector of after-state values squashed between -1 and 1. Each convolution layer has 2 types of filters \cite{young2016neurohex} : A $1 \times 1$ size filter initialized randomly and a 2d filter of $3 \times 3$ with last 2 rows of $1^{st}$ column, the entire $2^{nd}$ column and $1^{st}$ 2 rows of last column initialized randomly depicting the 6 neighbouring cells of a centre Hex cell with 2 remaining cells initialized to zero. The filter width is set to 6. The first layer has 2 filters of type I and 3 filters of type II. The second layer has 3 filters of type I and 2 filters of type II. The third layer has only 5 filters of type I. The fourth layer has a $245 \times 9 $ matrix and $9 \times 1$ vector initialized randomly.\par
The basic algorithm followed is the following \cite{sutton1998reinforcement}:

\begin{enumerate}
    \item Start at a random state $s_0$.
    \bindent
    \item process state through the neural network to choose an action $a$ $\epsilon$-greedily. Get $\hat{v}(s_t,w)$
    \item play action on the state.
    \item get reward $R_{t+1}$ and next state $s_{t+1}$.
    \item process the new state through the neural network to choose an action $b$ $\epsilon$-greedily. Get $\hat{v}(s_{t+1},w)$
    \item Update w as a function of $R_{t+1} + \hat{v}(s_{t+1},w) - \hat{v}(s_t,w)$
    \eindent
\end{enumerate}

The game is a 2-person game, however we do not train 2 networks representing 2 players. Instead we just train for one player, and assume that second player at best can copy the same strategy. This is sensible in such a game where the $1^{st}$ player has the winning strategy and if the first player makes a random choice in the middle, the second player can continue to play the $1^{st}$ player strategy and win. The self-play is a bit different than what one might use in zero-sum games.\par
\begin{figure}
    \centering
    \includegraphics[scale = 0.24]{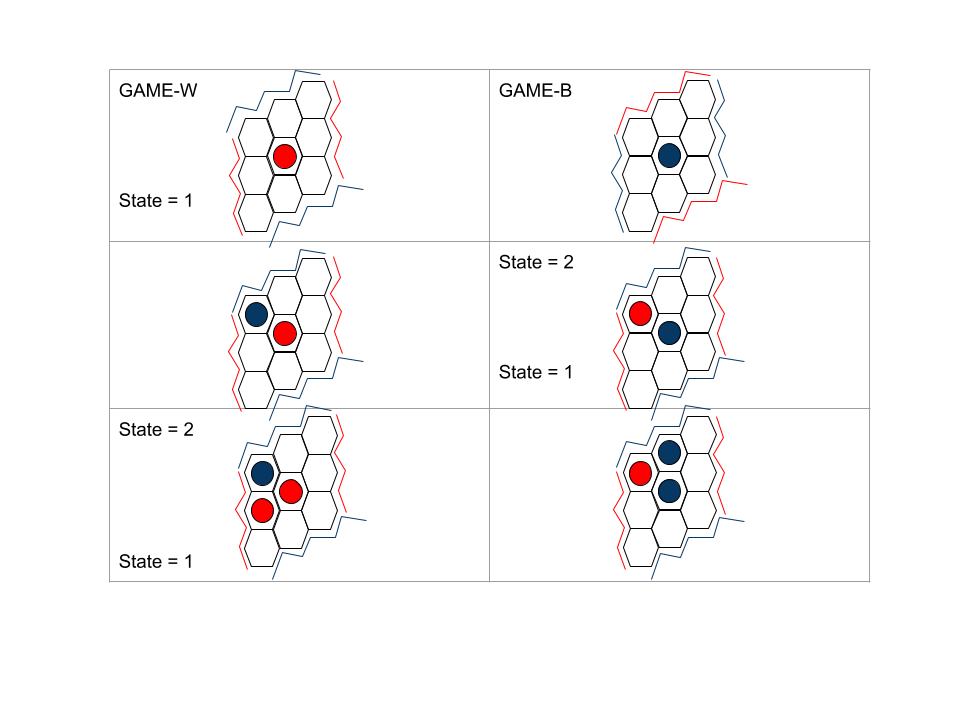}
    \caption{Self-Play as used in algorithm}
    \label{fig:Self Play}
\end{figure}

We make 2 copies of a board, in which one is the transpose of the other. In one copy, white moves first while in the other black moves first. The moves are mirror images of each other. Which one is to start at first is chosen at random. Once it's chosen, games are played on both copies of the board with maintaining the mirror image property of the moves. Two sequential states belong to different copies of the board (see figure \ref{fig:Self Play}). The mirror property of both copies and the use of convolution maintains that positions are evaluated equally on both copies of the game. This methodology is in the spirit of Hex, where the first person always has the winning strategy and by making the second player copy it ensures that the best policy learned will be used by both players to the same advantage. Another advantage is that this method takes into account the symmetricity of the Hex board, and thus ensures further exploration as well as exploitation \cite{young2016neurohex}. \par

\begin{figure}
    \centering
    \includegraphics[scale = 0.26]{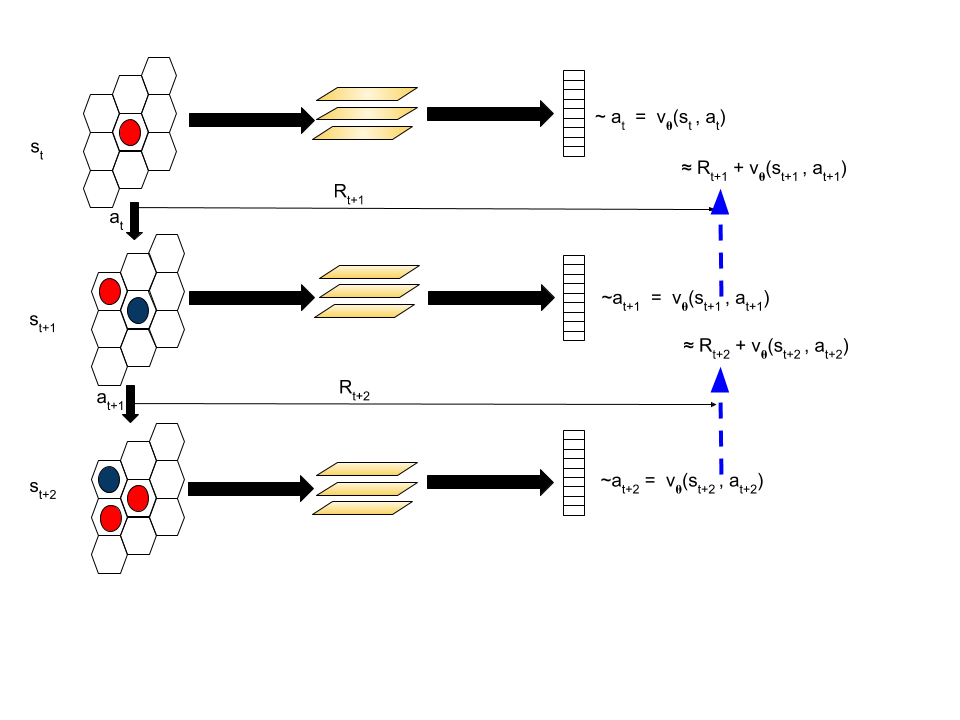}
    \caption{Flow-Model of training Network Parameters along with self play}
    \label{fig:Network 1}
\end{figure}

In Figure \ref{fig:Network 1} we show the basic algorithm in a pipeline form to represent error flow and self play simultaneously. As it is evident we have used online training of network, by updating parameters with every transition. However, this sort of training could result in high variance in parameter estimation, as the network parameters can get too fine tuned with a few states which are encountered too often. We therefore want to use batch training, where we train the network parameters simultaneously for a number of inputs \cite{young2016neurohex}. 

The updates, as shown in the figure, correspond to those of TD(0) \cite{AnAnalysisofTemporalDifferenceLearning} : \par
$\theta := \theta + \alpha[R_{t+1} + \hat{v}(s_{t+1}, b; \theta) - \hat{v}(s_t,\theta)]\nabla\hat{v}(s_t, a; \theta)$.
In literature this is defined as SARSA (state-action-reward-state-action) and is an on-policy TD-method.

This update based on reducing the Value Error $\mathbf{E}_\pi\big [ v_\pi(s) - \hat{v}(s,\theta) \big ]^2$ using bootstrap estimates of $v_\pi$.\par

We have also used Q-learning, which is an off-policy based TD-control procedure with the update rule \par 
$$\theta := \theta + \alpha[R_{t+1} + \max_{b \in \mathcal{A}(s)}\hat{v}(s_{t+1}, b; \theta) - \hat{v}(s_t, a; \theta)]\nabla_{\theta}\hat{v}(s_t, a; \theta)$$
based on reducing the error defined as above.
\begin{figure}
    \centering
    \includegraphics[scale = 0.26]{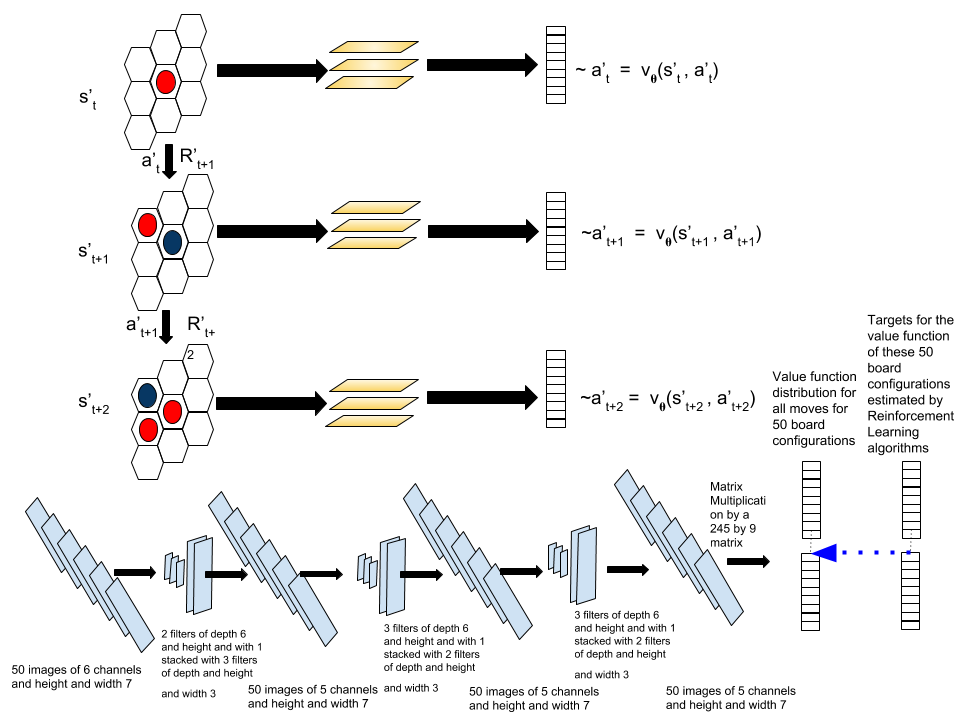}
    \caption{Flow-Model of training Network Parameters in Batch}
    \label{fig:Network 2}
\end{figure}
We have also used updates based on reducing the Projected Bellman Error $\mathbf{E}_\mu \big[ V_\theta - \mathbf{\Pi}T^\pi V_\theta \big]^2 $, where $\mathbf{\Pi}$ is a projection operator, $\mu$ the state distribution assumed under a behaviour policy $b$, $V_\theta$ the current value function estimate with parameters $\theta$ and $T^\pi$ the Bellman Operator acting on the value function with target policy $\pi$ \cite{NIPS20083626}. Based on the initial works of Sutton et al we had two prominent algorithms GTD 2 and TDC \cite{Sutton2009FGM15533741553501}. Their work primarily focused on linear function approximators where the projection operator could be represented as a matrix. Shalabh Bhatnagar et al also argued that under assumptions of small learning rate and smooth value-functions, the function-space represented by a neural network could be approximated to be linear around a small neighbourhood of the current parameter estimate. The authors then extended their work to general non-linear function approximators such as neural networks \cite{bhatnagar2009convergent}, \cite{maei2011gradient}.  The updates for GTD2 and TDC algorithm using non-linear function approximators are used for updating our network parameters. \par

\label{GTD 2 :}
\begin{algorithmic}
\STATE $\delta = R_{t+1} + \hat{v}(s_{t+1},b ; \theta) - \hat{v}(s_t,a ; \theta) $,
\STATE $w := \beta\Big(\delta - \nabla_\theta\hat{v}(s_t,a ; \theta)^Tw\Big)\nabla_\theta\hat{v}(s_t,a ; \theta)$,
\STATE $h = \Big(\delta - \nabla_\theta\hat{v}(s_t,a ; \theta)^Tw\Big)\nabla^2_\theta\hat{v}(s_t,a ; \theta)w$,
\STATE $\theta := \alpha\bigg\{\Big(\nabla_\theta\hat{v}(s_t,a ; \theta) - \nabla_\theta\hat{v}(s_{t+1},b;\theta)\Big)\Big(\nabla_\theta\hat{v}(s_t,a ;\theta)^Tw\Big)-h\bigg\}$.
\end{algorithmic}

\label{TDC  :}
\begin{algorithmic}
\STATE $\delta = R_{t+1} + \hat{v}(s_{t+1},b ; \theta) - \hat{v}(s_t,a ; \theta) $,
\STATE $w := \beta\Big(\delta - \nabla_\theta\hat{v}(s_t,a ; \theta)^Tw\Big)\nabla_\theta\hat{v}(s_t,a ; \theta)$,
\STATE $h = \Big(\delta - \nabla_\theta\hat{v}(s_t,a ; \theta)^Tw\Big)\nabla^2_\theta\hat{v}(s_t,a ; \theta)w$,
\STATE $\theta := \alpha\bigg\{\delta\nabla_\theta\hat{v}(s_t,a ; \theta) - \nabla_\theta\hat{v}(s_{t+1},b;\theta)\Big(\nabla_\theta\hat{v}(s_t,a ;\theta)^Tw\Big)-h\bigg\}$.
\end{algorithmic}

As one works through the derivation of the above algorithms one will notice that the starting point is the same \cite{NIPS20083626}, \cite{Sutton2009FGM15533741553501}, \cite{maei2011gradient}. One could conclude that the algorithms would behave almost similarly. Here $\alpha$ and $\beta$ are the two time scales. We have experimented by varying the time-scale ratios of both these algorithms. Hamid Maei and Richard Sutton have extended the TDC algorithm to include eligibility traces and even further to include prediction of action-values as well, which they call as the GQ($\lambda$) algorithm \cite{maei2010gq}. \textbf{By following the same procedure we have come up with the GTD2 algorithm which includes eligibility traces.} For the action-value prediction portion, we note that in our problem there is no distinct separation between state-value and action-value. We have used our algorithms in the setting of what Richard Sutton calls as after-state-values \cite{sutton1998reinforcement}. \par
Hamid Maei has also particularly extended the GQ($\lambda$) algorithm to control settings which they term as Greedy GQ($\lambda$) algorithm \cite{maei2011gradient}. This is along the same lines of using a random policy such as $\epsilon$-greedy as a behaviour policy which ensures exploration and a greedy policy with respect to the current estimate of action values. \textbf{We used the same ideas in our eligibility trace extended version of the GTD2 algorithm to extend to control problems.}  For completeness we have also implemented the Greedy GQ($\lambda$) algorithm separately though we assume its performance would be similar to that of the TDC algorithm used alongside an $\epsilon$-greedy policy as used in off-policy control. Our extensions are not documented because it is similar to the extensions carried out for the TDC algorithm. \par
We have implemented the three algorithms along with SARSA which includes eligibility traces as well and Q-learning in the procedural architecture depicted in Fig 2. That is, to update network parameters manually using update equations following single state transitions. However it is well known in the Deep-Learning Literature that single input-output training of neural networks might result in high variance in the parameter updates. So we want to use a batch of inputs and outputs to train our neural network. In THEANO, neural network training is done by giving a loss function, for example $\big[v_\pi(s) - \hat{v}(s,w)\big]^2$ and the network parameters are then updated to minimize the loss using a variety of optimization techniques. We have estimated $v_\pi(s)$ by Q-learning and have updated network-parameters after every 50 state transitions to optimize  $\big[v_\pi(s) - \hat{v}(s,w)\big]^2$ using a single step of RMSPROP \cite{tieleman2012lecture} for a 50 state-batch as depicted in Figure 3. However, minimizing the projected Bellman Error in this way presents a problem. The projection operator for curved general surfaces is difficult to depict and hence the loss function could not be represented with known parameters. \textbf{The way we have bypassed this problem is optimizing two objective functions simultaneously:}
$$\Pi Tv_{\hat{\theta}} = \min_{\theta}[ Tv_\theta - v_\theta ]^2, $$
$$\theta^* =  \argmin_{\theta}[\Pi Tv_{\hat{\theta}} - v_\theta]^2, $$
where $Tv_\theta$ is approximated by Q-learning. After 50 state transitions, we have an estimate of $Tv_\theta$ by Q-learning and we run a single step of RMSPROP to get an approximation of $\Pi Tv_{\hat{\theta}}$ followed by a single step of RMSPROP to update network parameters to be used again in the next 50 state-transitions. Though we have not proved that this will be equivalent to minimizing the Projected Bellman Error, we shall nevertheless call it as the PBE.\par
Apart from using $\epsilon$-greedy policy to control we have also used policy-gradient techniques to explicitly optimize for an optimal policy. Specifically we used actor-critic algorithms, where a separate value or critic network is used to estimate the value function and a policy or actor network is used to optimize the policy parameters. We have included eligibility traces during policy parameter updates as well \cite{sutton1998reinforcement}. The policy network is similar to the value network \cite{young2016neurohex} except that in the $4^{th}$ layer instead of a sigmoid function we have used a softmax function which acts on the product of a matrix and the output of the $3^{rd}$ layer. The general algorithm is suggested as follows:
\begin{enumerate}
    \item update parameters of value network.
    \item update parameters of policy network.
\end{enumerate}
We have used gradient ascent as well as Natural Gradient ascent \cite{bhatnagar2009natural, amari1998natural, pascanu2013revisiting} to update policy network parameters. We have used the setting of Fig. 2. that is updating parameters online and not used batch updates. The value function has been updated using SARSA, TDC, GTD 2, Greedy GQ($\lambda$) and Q-learning.\par 
To view how our present techniques do with older methodologies, we have resorted to use the H-search algorithm to score human played board moves \cite{young2016neurohex}. We constructed our own data-set of human moves on a $3 \times 3$ board and scored them via the H-search algorithm as discussed in Kenny Young et al. Using these scores we have pre-trained the neural network parameters, before using self-play and reinforcement learning algorithms (this methodology is used also in \cite{silver2016mastering}). We have compared these results with those where there had been no pre-training of neural networks and parameters were learned solely by self-play and reinforcement learning algorithms. This experiment was inspired by the recent success of Alpha Go Zero \cite{silver2017mastering}. Our results indicate no visible deviations with regards to the learned values and we conclude that self-play and reinforcement learning are sufficient for learning.\par
\begin{figure}
    \centering
    \includegraphics[scale = 0.26]{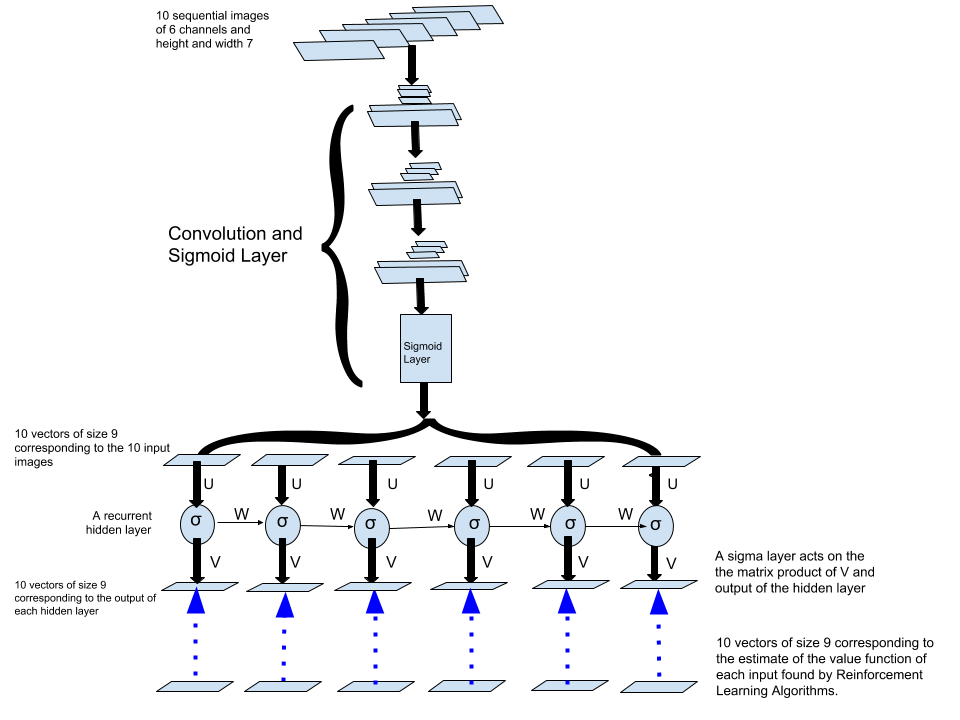}
    \caption{Flow-Model of training a Recurrent Neural Network in batch.}
    \label{fig:Network 3}
\end{figure}

Finally, we have experimented with a separate architecture itself. The first 4 layers are similar to the old architecture on top of which we have applied a recurrent neural network of depth 10. A sequential batch of 10 board configurations is fed through the neural network. The output of the first 3 convolution layers followed by the sigmoid layer( See Fig \ref{fig:Network 2}) results in 10 vectors of size 9. We then apply the Recurrent Layer as shown in Fig. \ref{fig:Network 3}. Here, U,V and W are three $9 \times 9$ matrices. As activation function we use a sigmoid function. Our intuition is, by making a board configuration at a later stage dependent on board configurations encountered earlier, one might get improvements. The target updates for each input are estimated using Q-learning as well as SARSA. The loss functions that the THEANO environment tries to minimize are both the Mean Squared Error and the Projected Bellman Error between the target and the current estimate. We use the RMSPROP algorithm to optimize.

\section{Results}
We plot after-state-values averaged over each trajectory as a function of the number of iterations. Since a reward of 1 is obtained only upon termination after a win and 0 everywhere else, one could quickly see that optimal after-state values would be 1.\par
\begin{figure}
\centering
\subfigure[]{\label{fig : 5a}\includegraphics[scale=0.15]{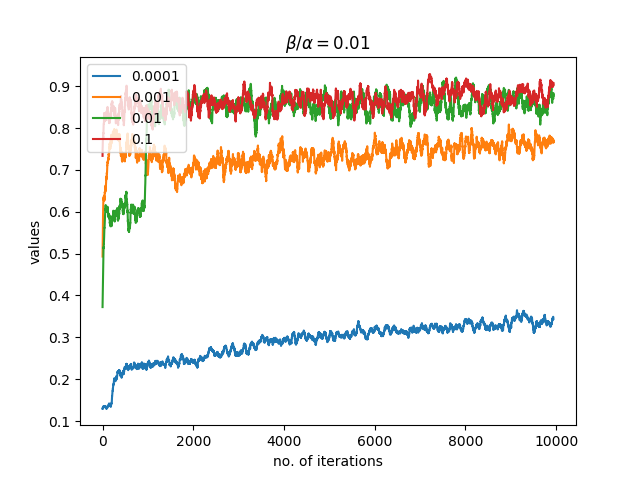}}\hspace{0pt}
\subfigure[]{\label{fig : 5b}\includegraphics[scale=0.15]{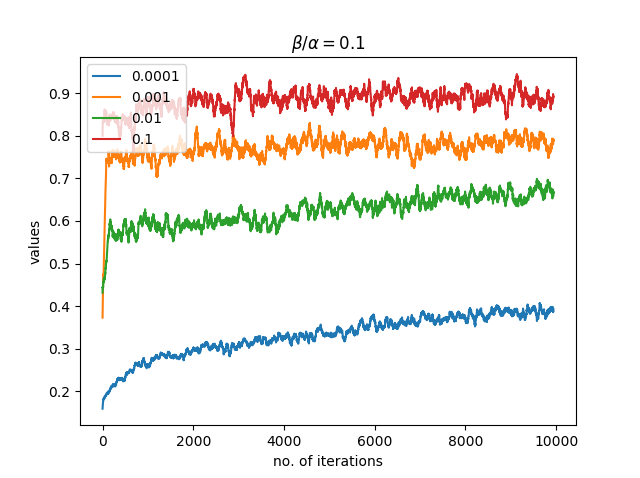}}
\subfigure[]{\label{fig : 5c}\includegraphics[scale=0.15]{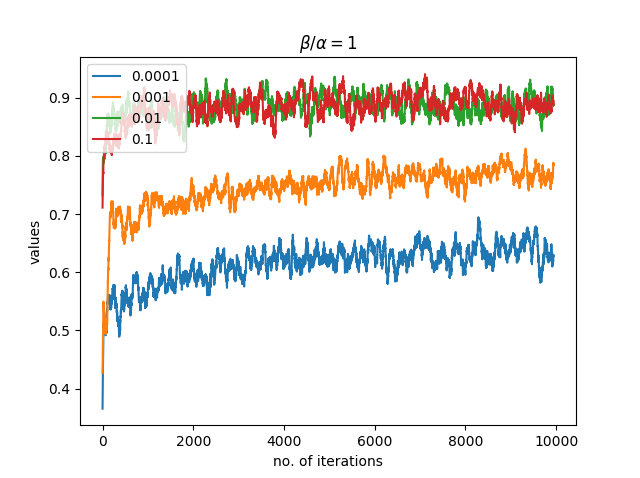}}\hspace{0pt}
\subfigure[]{\label{fig : 5d}\includegraphics[scale=0.15]{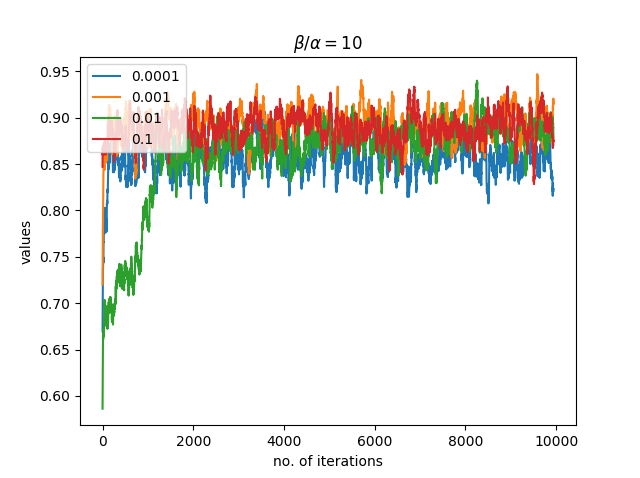}}
\subfigure[]{\label{fig : 5e}\includegraphics[scale=0.15]{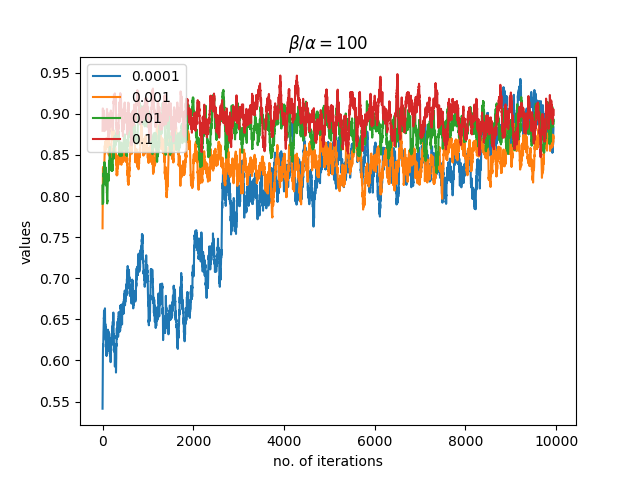}}\hspace{0pt}
\caption[results of GTD 2 algorithm]{After state values using GTD 2 algorithm for $\alpha$ values 0.0001, 0.001, 0.01 and 0.1 and $\beta/\alpha$ ratios at:
        \subref{fig : 5a} $\beta/\alpha = 0.01$
        \subref{fig : 5b} $\beta/\alpha = 0.1$
        \subref{fig : 5c} $\beta/\alpha = 1$
        \subref{fig : 5d} $\beta/\alpha = 10$
        \subref{fig : 5e} $\beta/\alpha = 100$}
\label{fig : GTD 2 }
\end{figure}
\begin{figure}
\centering
\subfigure[]{\label{fig : 6a}\includegraphics[scale=0.15]{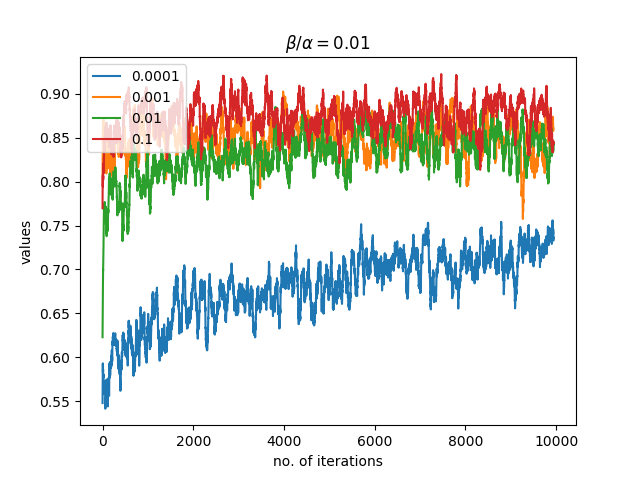}}\hspace{2pt}
\subfigure[]{\label{fig : 6b}\includegraphics[scale=0.15]{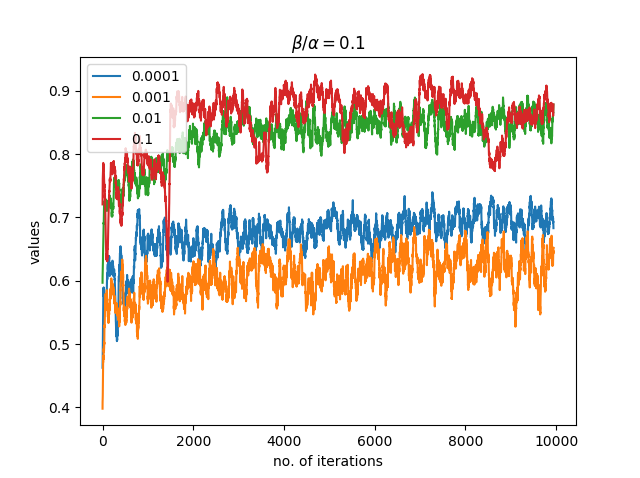}}
\subfigure[]{\label{fig : 6c}\includegraphics[scale=0.15]{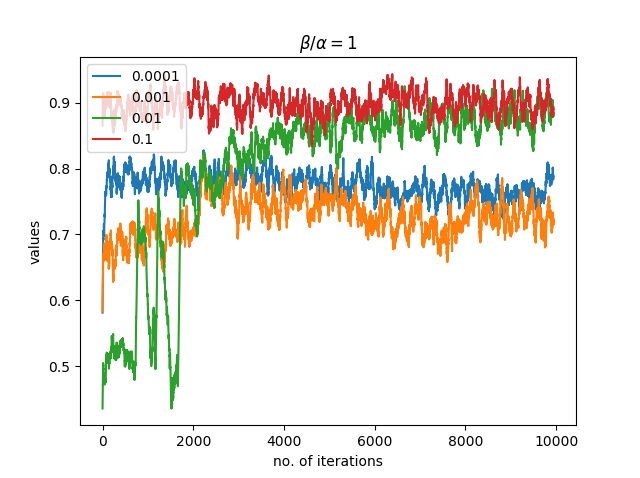}}\hspace{2pt}
\subfigure[]{\label{fig : 6d}\includegraphics[scale=0.15]{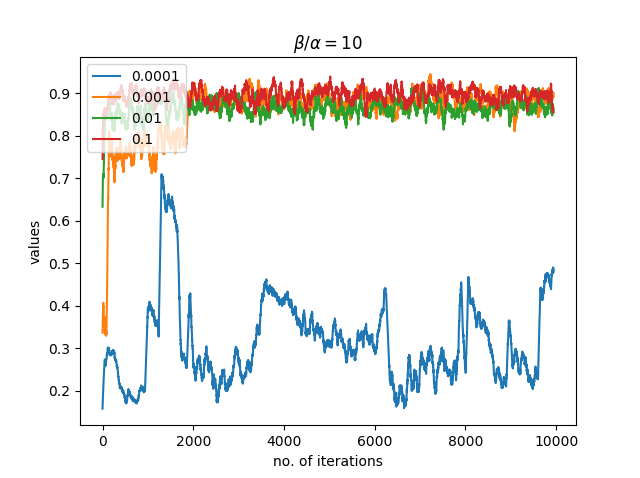}}
\subfigure[]{\label{fig : 6e}\includegraphics[scale=0.15]{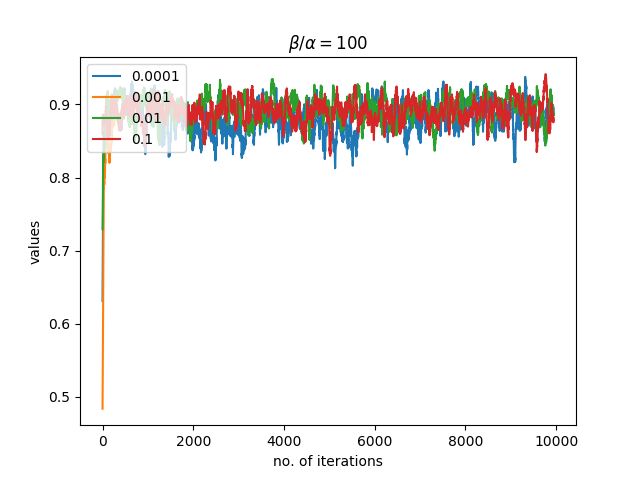}}\hspace{2pt}
\caption[results of TDC algorithm]{After state values using TDC algorithm for $\alpha$ values 0.0001, 0.001, 0.01 and 0.1 and $\beta/\alpha$ ratios at :
        \subref{fig : 6a} $\beta/\alpha = 0.01$
        \subref{fig : 6b} $\beta/\alpha = 0.1$
        \subref{fig : 6c} $\beta/\alpha = 1$
        \subref{fig : 6d} $\beta/\alpha = 10$
        \subref{fig : 6e} $\beta/\alpha = 100$}
\label{fig : TDC }
\end{figure}
\begin{figure}
\centering
\subfigure[]{\label{fig : 7a}\includegraphics[scale=0.15]{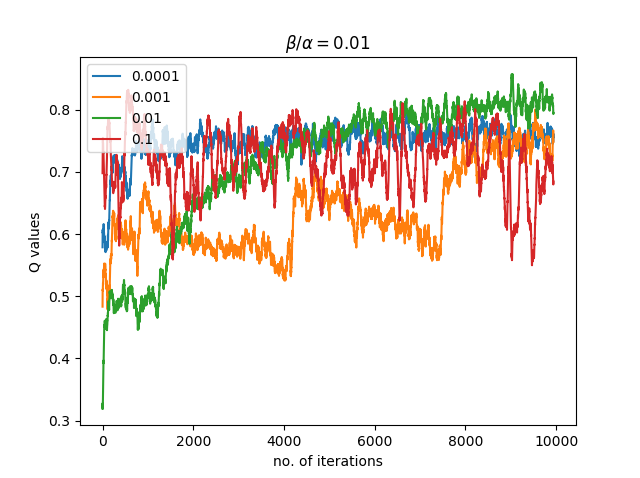}}\hspace{2pt}
\subfigure[]{\label{fig : 7b}\includegraphics[scale=0.15]{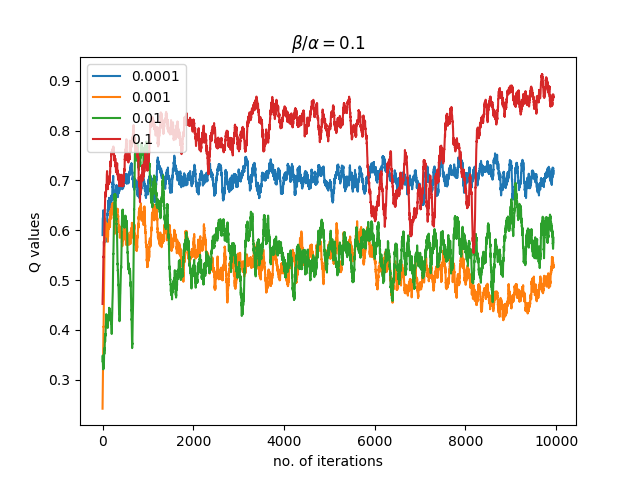}}
\subfigure[]{\label{fig : 7c}\includegraphics[scale=0.15]{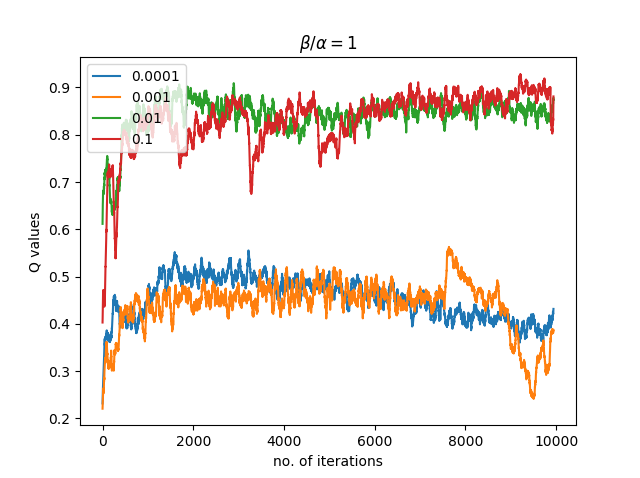}}\hspace{2pt}
\subfigure[]{\label{fig : 7d}\includegraphics[scale=0.15]{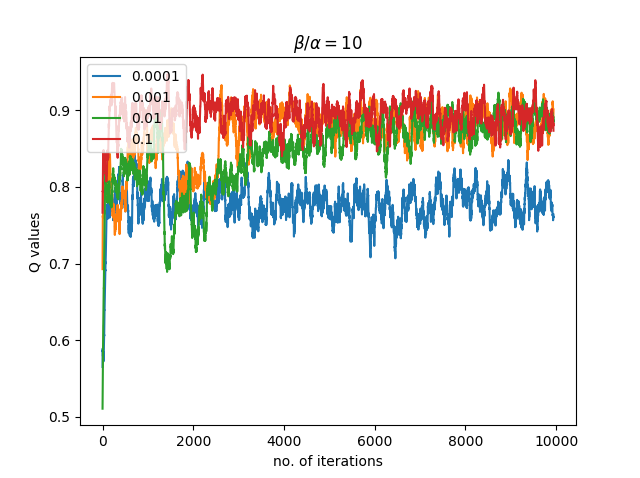}}
\subfigure[]{\label{fig : 7e}\includegraphics[scale=0.15]{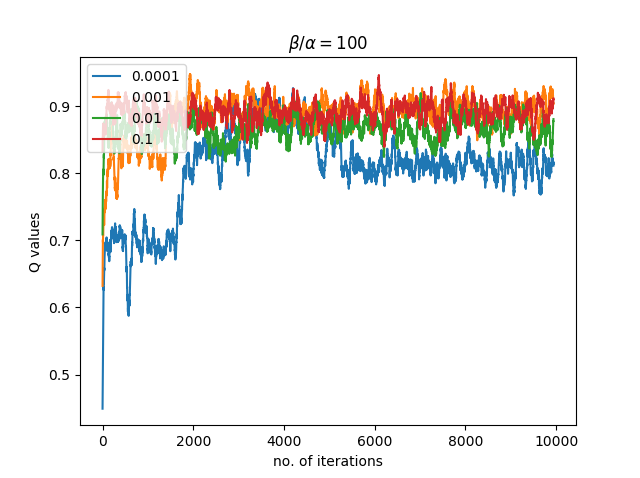}}\hspace{2pt}
\caption[results of the Greedy GQ algorithm]{After state values using Greedy GQ($\lambda$ algorithm for $\alpha$ values 0.0001, 0.001, 0.01 and 0.1 and $\beta/\alpha$ ratios at :
        \subref{fig : 7a} $\beta/\alpha = 0.01$
        \subref{fig : 7b} $\beta/\alpha = 0.1$
        \subref{fig : 7c} $\beta/\alpha = 1$
        \subref{fig : 7d} $\beta/\alpha = 10$
        \subref{fig : 7e} $\beta/\alpha = 100$}
\label{fig : GQ }
\end{figure}
In Figures \ref{fig : GTD 2 }, \ref{fig : TDC } and \ref{fig : GQ } we plot the after-state values for 5 two-time scale ratios and 4 value-network parameter learning rates $\alpha$ in a network shown in Fig. \ref{fig:Network 1} using the GTD2, TDC and Greedy GQ($\lambda$) update algorithms following online implementation. \par
\begin{figure}
\centering
\subfigure[]{\label{fig : 8a}\includegraphics[scale=0.15]{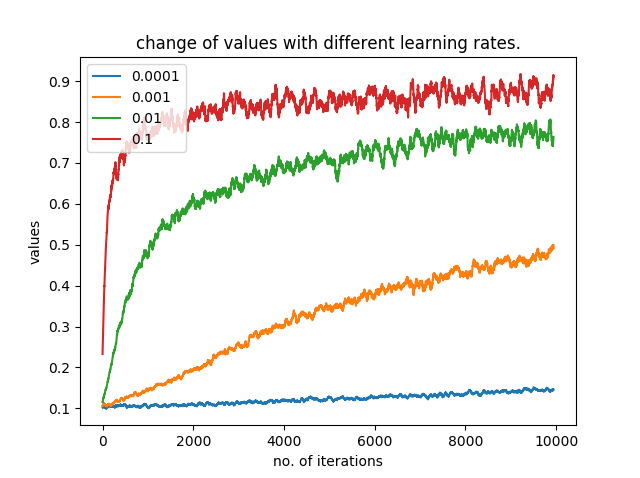}}\hspace{2pt}
\subfigure[]{\label{fig : 8b}\includegraphics[scale=0.15]{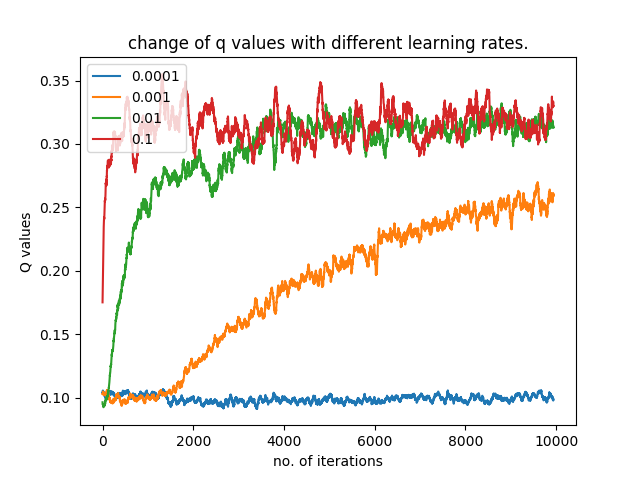}}
\caption[results of TD and Q algorithm]{
        \subref{fig : 8a} After state values using SARSA algorithm for $\alpha$ values 0.0001, 0.001, 0.01 and 0.1
        \subref{fig : 8b} After state values using Q learning algorithm for $\alpha$ values 0.0001, 0.001, 0.01 and 0.1}
\label{fig : TD and Q }
\end{figure}
In  Fig. \ref{fig : TD and Q } we plot the after-state values for 4 learning rates of the value-network parameter where the network is the one shown in Fig \ref{fig:Network 1} using SARSA and Q-learning updates of targets implemented online.\par
\begin{figure}
\centering
\subfigure[]{\label{fig : 9a}\includegraphics[scale=0.15]{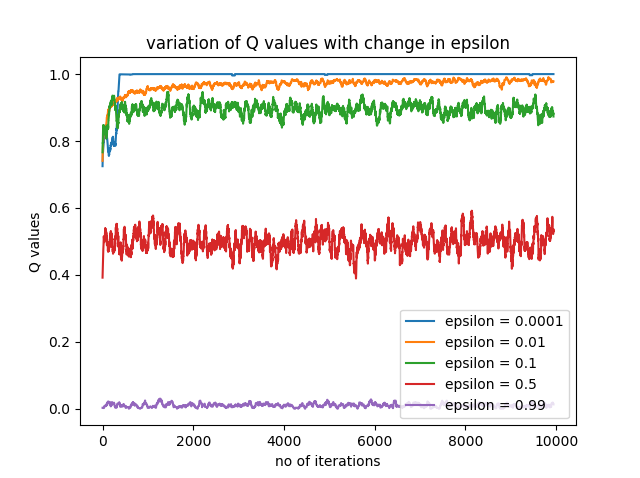}}\hspace{2pt}
\subfigure[]{\label{fig : 9b}\includegraphics[scale=0.15]{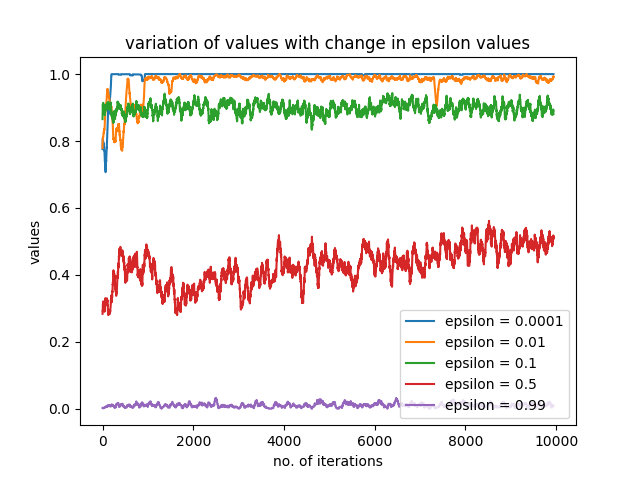}}
\subfigure[]{\label{fig : 9c}\includegraphics[scale=0.15]{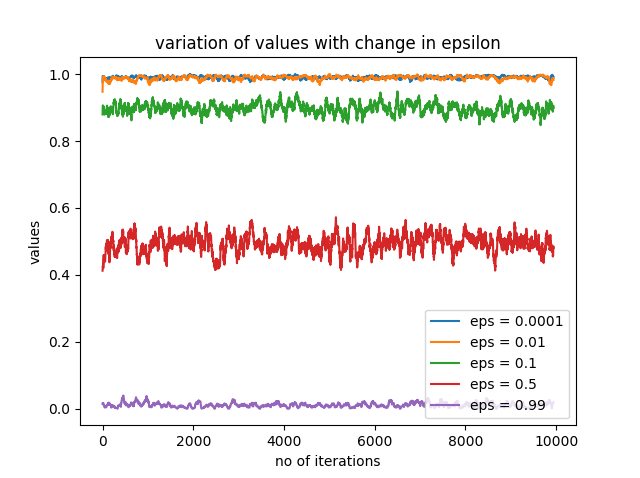}}\hspace{2pt}
\subfigure[]{\label{fig : 9d}\includegraphics[scale=0.15]{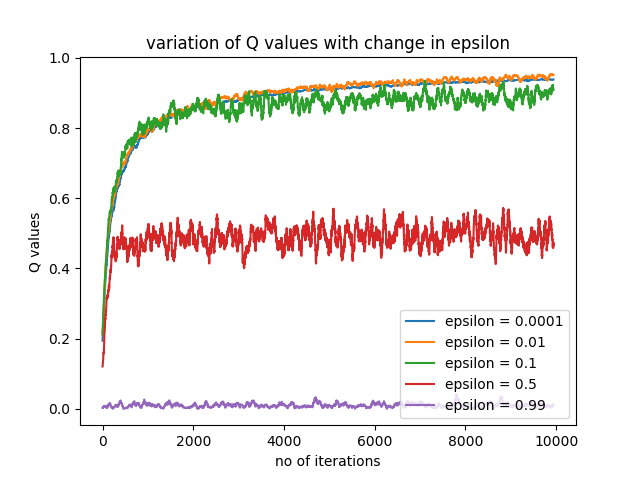}}
\subfigure[]{\label{fig : 9e}\includegraphics[scale=0.15]{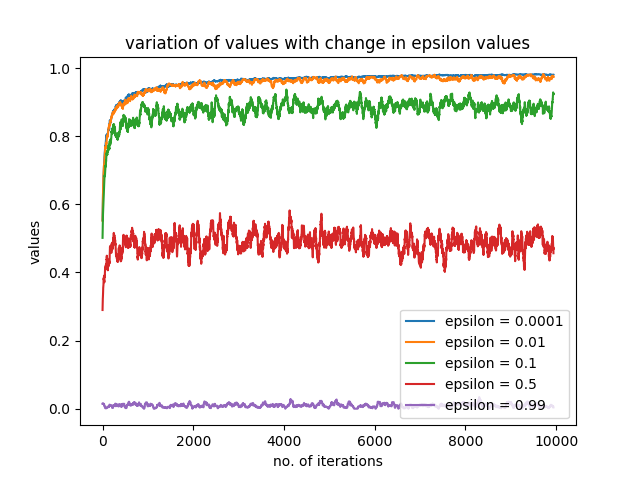}}\hspace{2pt}
\caption[results of changing $\epsilon$ values]{After state values with $\epsilon$ values at 0.0001, 0.01, 0.1, 0.5 and 0.99 for:
        \subref{fig : 9a} Greedy GQ$\lambda$ algorithm
        \subref{fig : 9b} TDC algorithm
        \subref{fig : 9c} GTD 2 algorithm
        \subref{fig : 9d} Q-learning algorithm
        \subref{fig : 9e} SARSA algorithm}
\label{fig : epsilon-values }
\end{figure}

\begin{figure}
\centering
\subfigure[]{\label{fig : 10a}\includegraphics[scale=0.15]{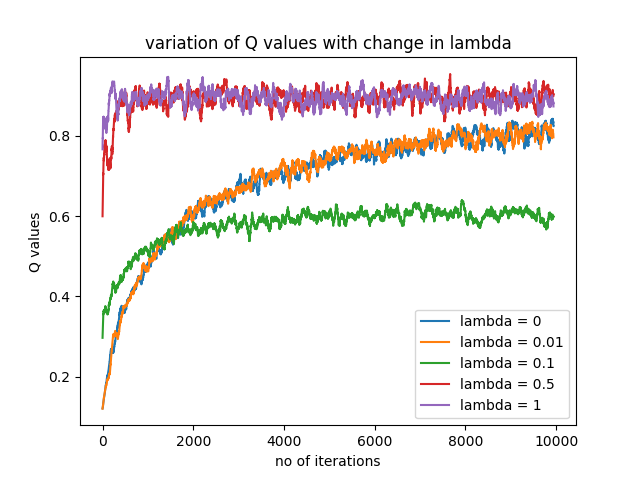}}\hspace{2pt}
\subfigure[]{\label{fig : 10b}\includegraphics[scale=0.15]{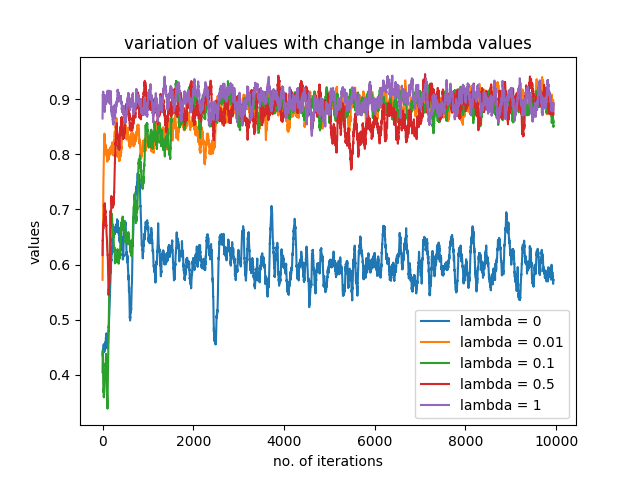}}
\subfigure[]{\label{fig : 10c}\includegraphics[scale=0.15]{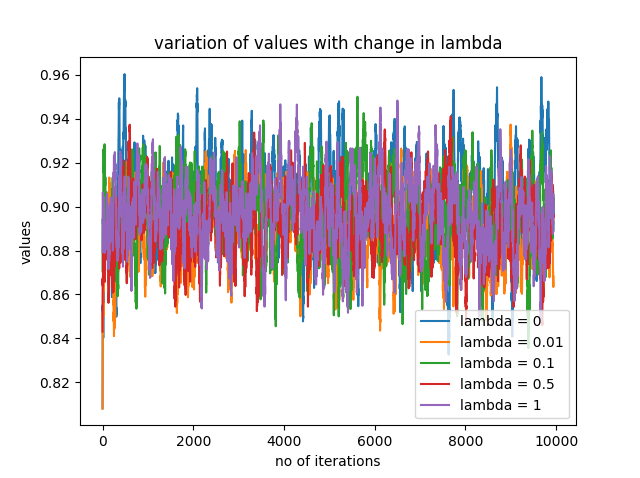}}\hspace{2pt}
\subfigure[]{\label{fig : 10d}\includegraphics[scale=0.15]{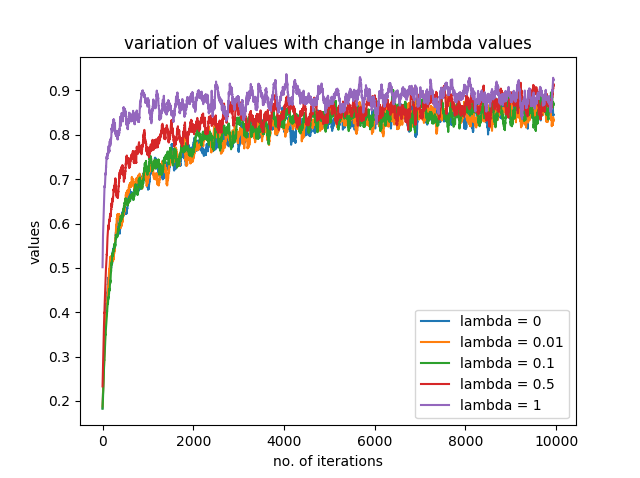}}\hspace{2pt}
\caption[results of changing $\lambda$ values]{After State Values with $\lambda$ values at 0, 0.01, 0.1, 0.5 and 1 for:
        \subref{fig : 9a} Greedy GQ$\lambda$ algorithm
        \subref{fig : 9b} TDC algorithm
        \subref{fig : 9c} GTD 2 algorithm
        \subref{fig : 9e} SARSA algorithm}
\label{fig : lambda-values }
\end{figure}
In Fig. \ref{fig : epsilon-values } we plot the after-state values for five $\epsilon$ values, the probability with which greedy actions are chosen over random actions, using value-network updates, in a network shown in Fig \ref{fig:Network 1} done online, given by the Greedy GQ($\lambda$), TDC, GTD2, Q-learning and SARSA algorithms respectively.\par
In Fig. \ref{fig : lambda-values } we show the variation of after-state values for 5 eligibility trace parameters $\lambda$ using updates given by Greedy GQ($\lambda$), TDC, GTD2, and SARSA algorithms in the network shown in Fig. \ref{fig:Network 1} and implemented online.\par
\begin{figure}
\centering
\subfigure[]{\label{fig : 11a}\includegraphics[scale=0.15]{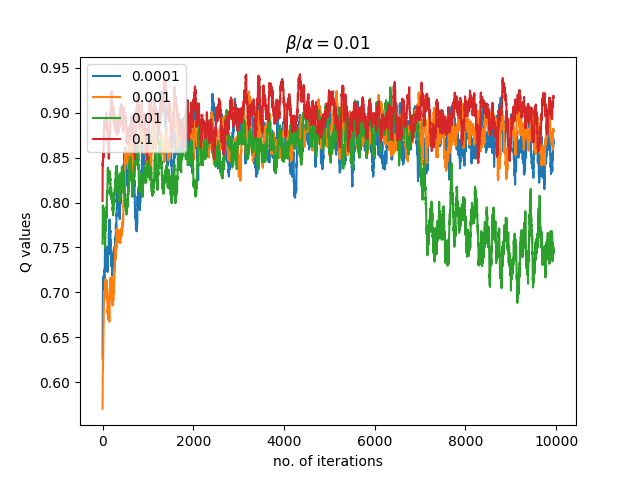}}\hspace{2pt}
\subfigure[]{\label{fig : 11b}\includegraphics[scale=0.15]{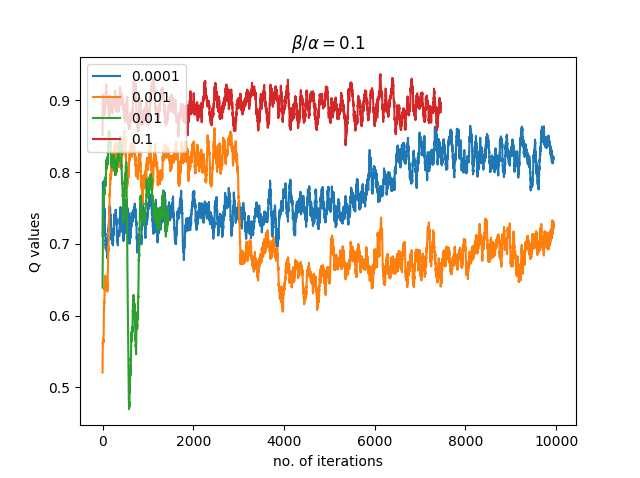}}
\subfigure[]{\label{fig : 11c}\includegraphics[scale=0.15]{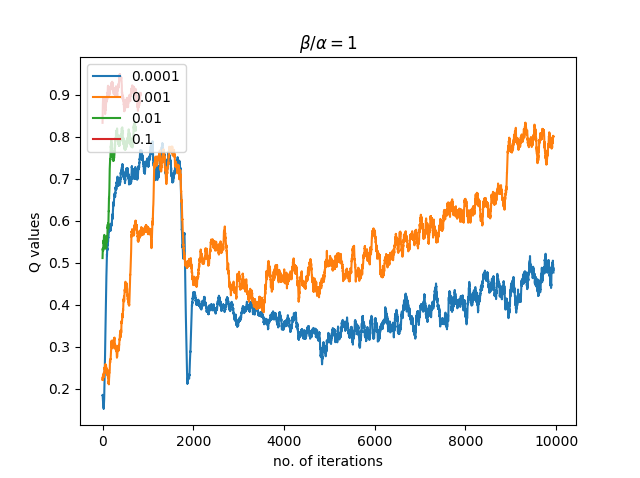}}\hspace{2pt}
\subfigure[]{\label{fig : 11d}\includegraphics[scale=0.15]{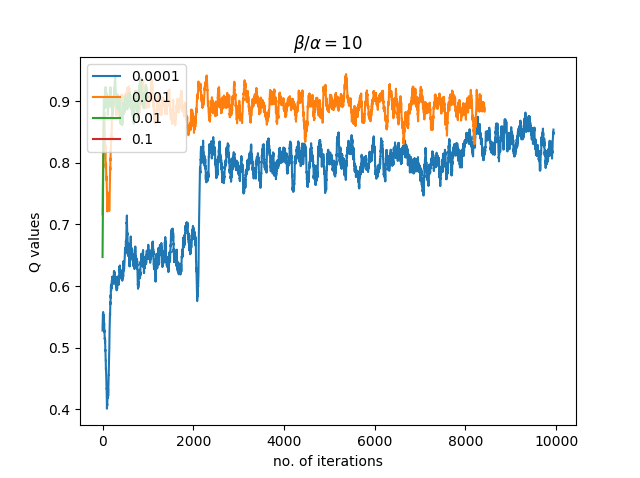}}
\subfigure[]{\label{fig : 11e}\includegraphics[scale=0.15]{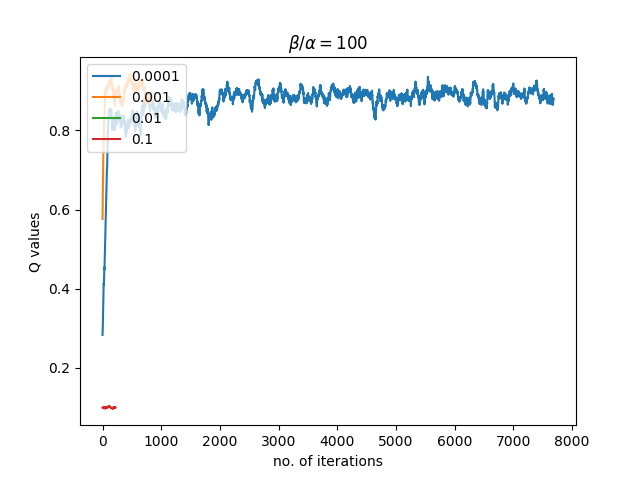}}\hspace{2pt}
\caption[results of NGTDC algorithm]{After state values using TDC algorithm with Natural Gradient for optimizing value-network parameters with $\alpha$ values 0.0001, 0.001, 0.01 and 0.1 and $\beta/\alpha$ ratios at:
        \subref{fig : 11a} $\beta/\alpha = 0.01$
        \subref{fig : 11b} $\beta/\alpha = 0.1$
        \subref{fig : 11c} $\beta/\alpha = 1$
        \subref{fig : 11d} $\beta/\alpha = 10$
        \subref{fig : 11e} $\beta/\alpha = 100$}
\label{fig : NGTDC }
\end{figure}
In Fig. \ref{fig : NGTDC } we plot after-state values for a network, shown in figure \ref{fig:Network 1},  where the parameters have been updated online using the TDC algorithm, but instead of using stochastic gradient to optimize we use Natural Gradients. The results are shown for 5 different ratios of the 2 time-scales and 5 different learning rates. We call this method as TDC with NG.\par 

\begin{figure}
\centering
\subfigure[]{\label{fig : 12a}\includegraphics[scale=0.15]{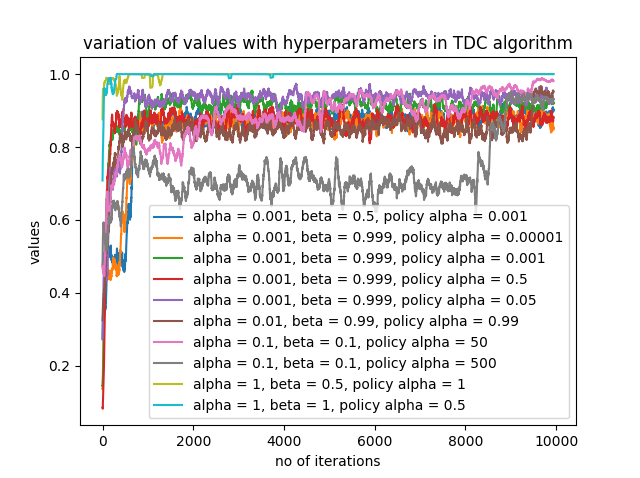}}\hspace{2pt}
\subfigure[]{\label{fig : 12b}\includegraphics[scale=0.15]{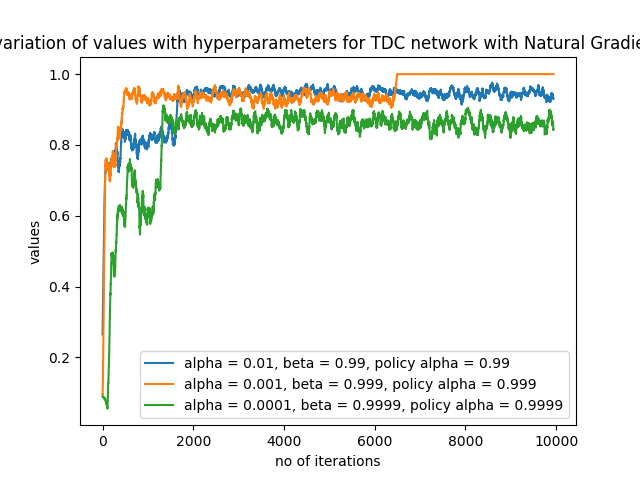}}
\subfigure[]{\label{fig : 12c}\includegraphics[scale=0.15]{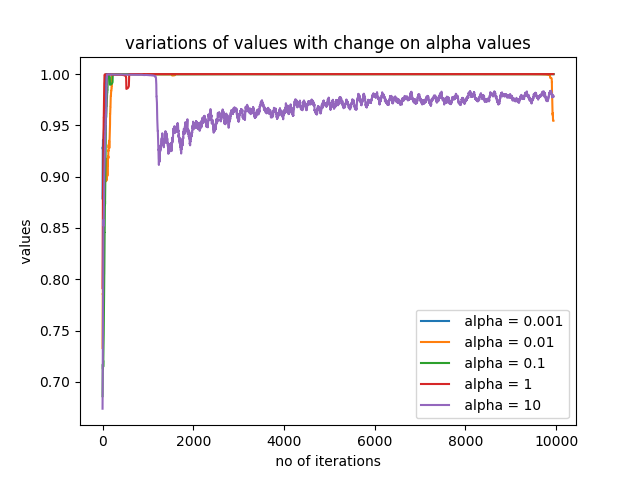}}\hspace{2pt}
\subfigure[]{\label{fig : 12d}\includegraphics[scale=0.15]{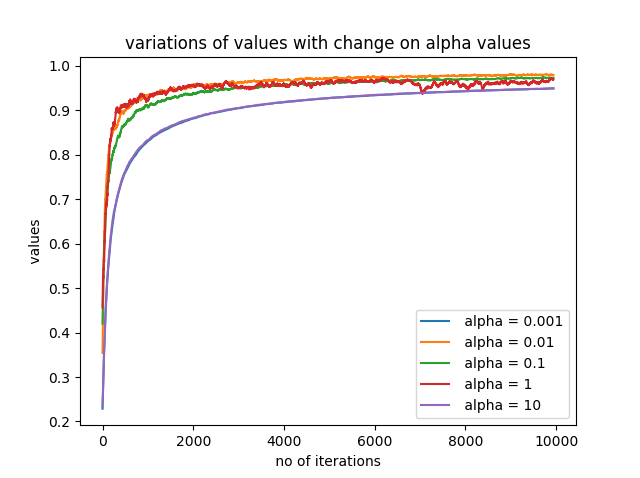}}
\subfigure[]{\label{fig : 12e}\includegraphics[scale=0.15]{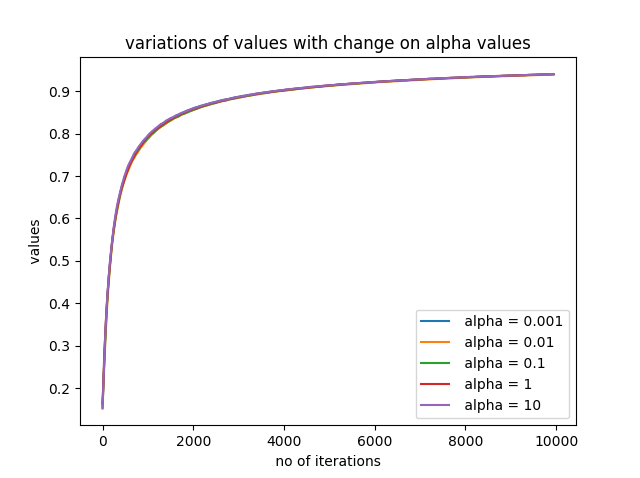}}\hspace{2pt}
\caption[results of using policy gradient]{After-State Values using Policy Gradient to optimize a separate policy network with learning rates $\alpha$ at 0.001, 0.01, 0.1, 1 and 10 for :
        \subref{fig : 12a} TDC algorithm
        \subref{fig : 12b} TDC algorithm with Natural Gradient optimization of policy network
        \subref{fig : 12c} GTD 2 algorithm
        \subref{fig : 12d} Greedy GQ($\lambda$) algorithm
        \subref{fig : 12e} Q-learning algorithm.}
\label{fig : Policy-Gradient }
\end{figure}
In Fig. \ref{fig : Policy-Gradient } we plot after-state values generated while using a separate policy network to learn an explicit policy.The policy network is trained using stochastic gradient and results are shown for five learning rates $\alpha$ with value-network updated using TDC, GTD2, Greedy GQ($\lambda$) and Q-learning. We also have one figure where Natural Gradient optimization is used for updating Policy Network parameters while the Value Function is updated using the TDC algorithm. The network used is shown in Fig. \ref{fig:Network 1} and updates are done online.\par
\begin{figure}
\centering
\subfigure[]{\label{fig : 13a}\includegraphics[scale=0.15]{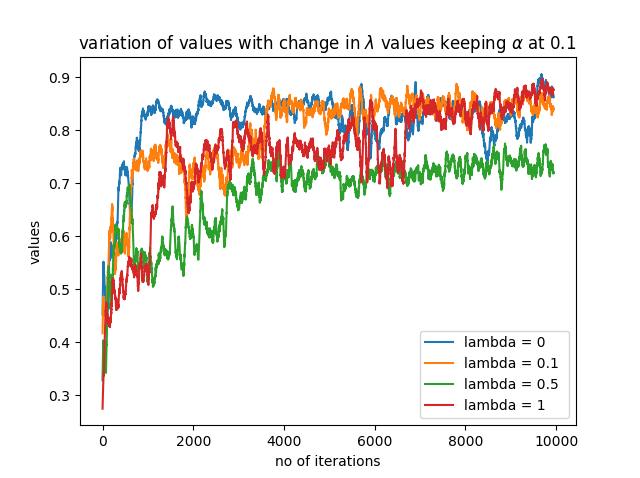}}\hspace{2pt}
\subfigure[]{\label{fig : 13b}\includegraphics[scale=0.15]{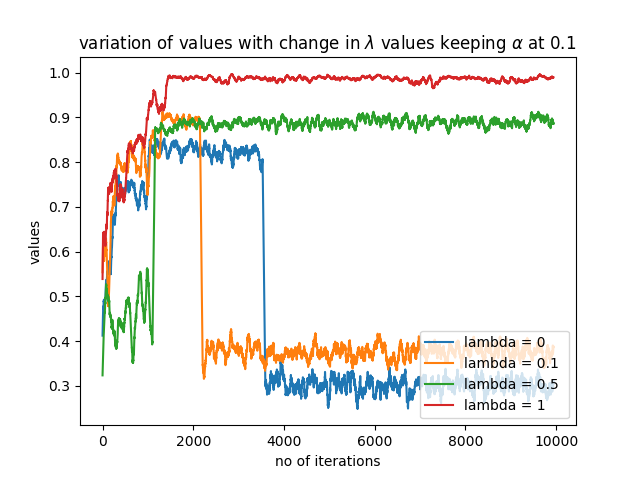}}
\subfigure[]{\label{fig : 13c}\includegraphics[scale=0.15]{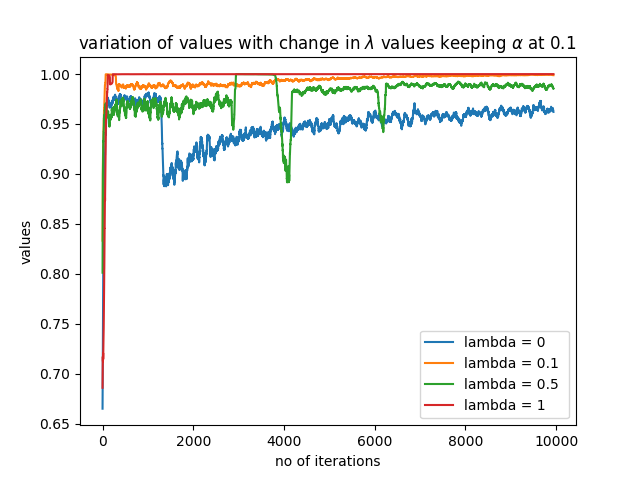}}\hspace{2pt}
\subfigure[]{\label{fig : 13d}\includegraphics[scale=0.15]{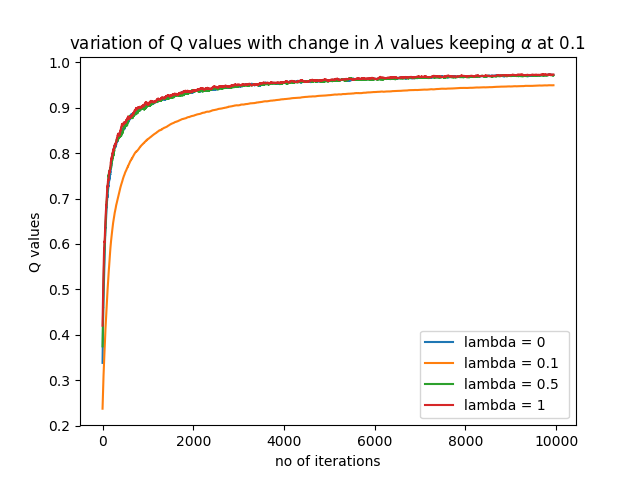}}
\subfigure[]{\label{fig : 13e}\includegraphics[scale=0.15]{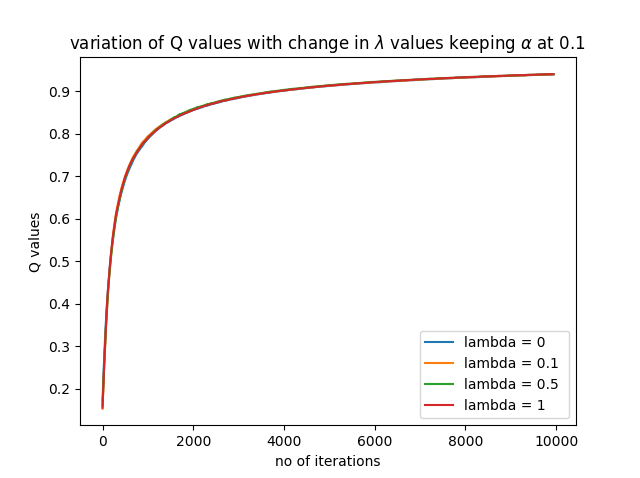}}\hspace{2pt}
\caption[results of using policy gradient with varying $\lambda$]{After-state values with policy parameters using eligibility traces $\lambda$ at 0, 0.1, 0.5 and 1 for :
        \subref{fig : 13a} TDC algorithm
        \subref{fig : 13b} TDC algorithm with Natural Gradient optimization of policy network
        \subref{fig : 13c} GTD 2 algorithm
        \subref{fig : 13d} Greedy GQ($\lambda$) algorithm
        \subref{fig : 13e} Q-learning algorithm.}
\label{fig : Policy-Gradient-lambda }
\end{figure}
In Fig. \ref{fig : Policy-Gradient-lambda } we plot after-state values when a separate policy network has been used. The policy parameters were updated in a network depicted in Fig. \ref{fig:Network 1} implemented online, by stochastic gradient using five eligibility traces $\lambda$ in the policy parameters themselves. Results are shown for each case when the value network is updated using TDC, GTD2, Greedy GQ($\lambda$) and Q-learning. We also present a figure when the policy parameters were updated using Natural Gradient instead of Stochastic Gradient. \par

\begin{figure}
\centering
\subfigure[]{\label{fig : 14a}\includegraphics[scale=0.15]{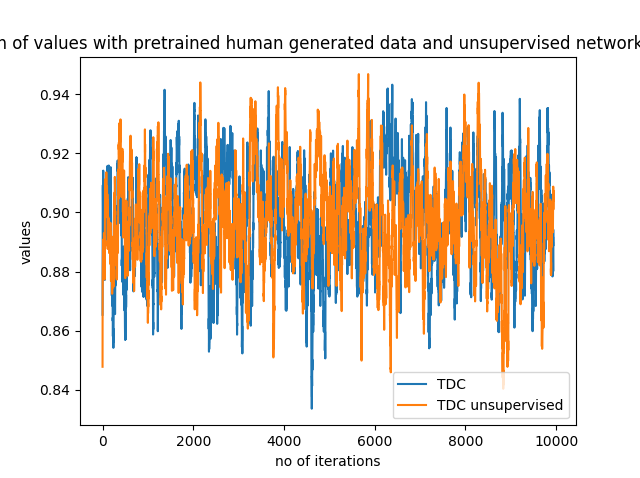}}\hspace{2pt}
\subfigure[]{\label{fig : 14b}\includegraphics[scale=0.15]{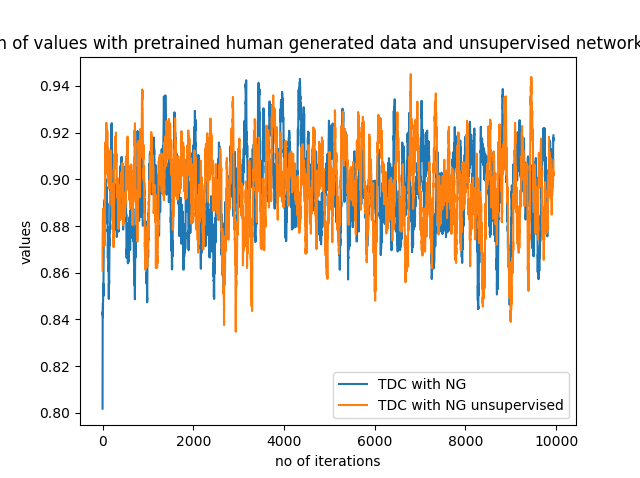}}
\subfigure[]{\label{fig : 14c}\includegraphics[scale=0.15]{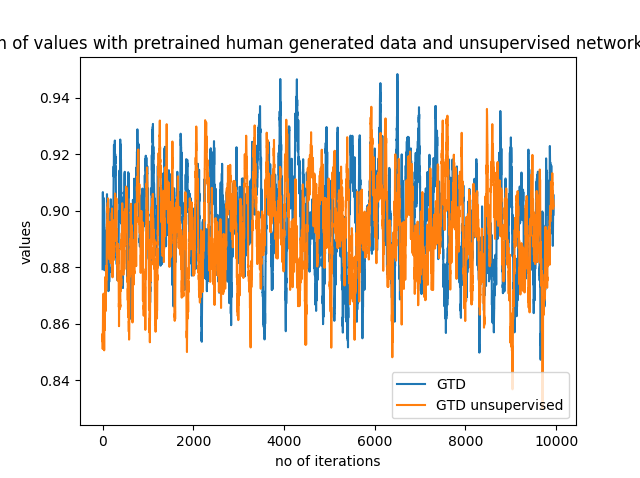}}\hspace{2pt}
\subfigure[]{\label{fig : 14d}\includegraphics[scale=0.15]{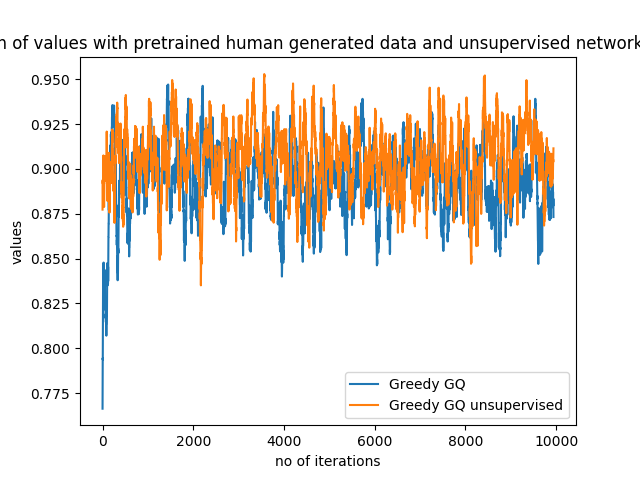}}
\subfigure[]{\label{fig : 14e}\includegraphics[scale=0.15]{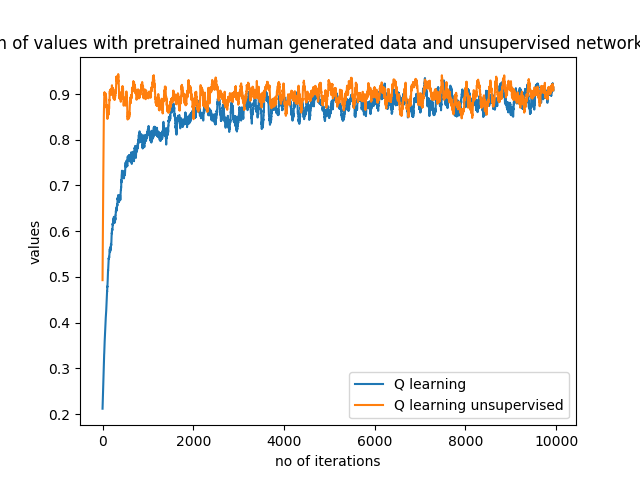}}\hspace{2pt}
\subfigure[]{\label{fig : 14f}\includegraphics[scale=0.15]{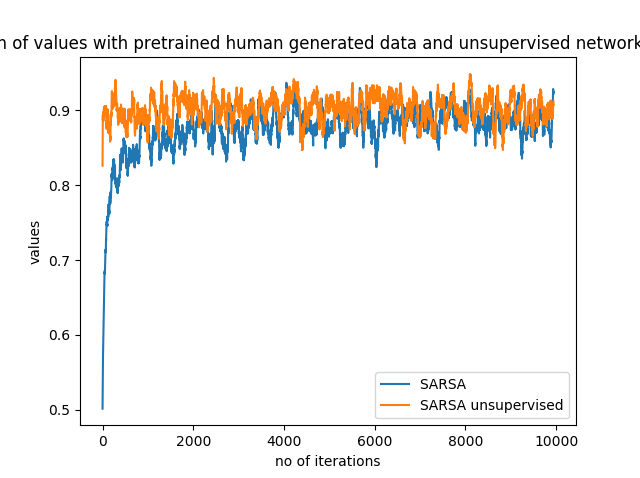}}
\caption[comparing between supervised and unsupervised]{After state values with and without pretraining using human dataset :
        \subref{fig : 14a} TDC algorithm
        \subref{fig : 14b} TDC algorithm with Natural Gradient optimization of policy network
        \subref{fig : 14c} GTD 2 algorithm
        \subref{fig : 14d} Greedy GQ($\lambda$) algorithm
        \subref{fig : 14e} Q-learning algorithm.
        \subref{fig : 14f} SARSA algorithm.}
\label{fig : Compare }
\end{figure}
In Fig. \ref{fig : Compare } we compare the after state values with a supervised pre-training of network (Fig. \ref{fig:Network 1}) with parameters updated online, using constructed data of played games and scored according to the idea of H-search algorithm followed by self-play and reinforcement learning of network parameters, without any human data relying only on self play and reinforcement learning algorithms. \par
\begin{figure}
\centering
\subfigure[]{\label{fig : 15a}\includegraphics[scale=0.15]{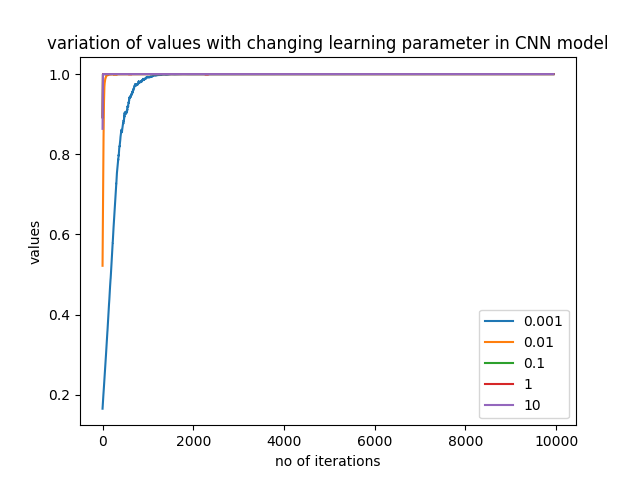}}\hspace{2pt}
\subfigure[]{\label{fig : 15b}\includegraphics[scale=0.15]{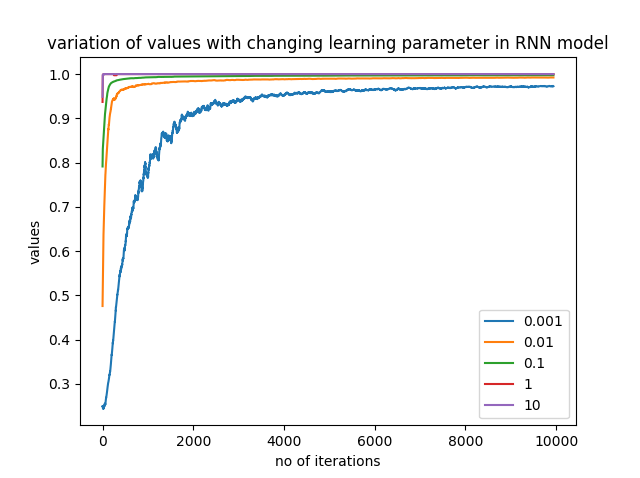}}
\caption[comparison of CNN and RNN model]{After state values implemented using Q-learning and batch updates for 
        \subref{fig : 15a} CNN model with learning rates $\alpha$ at 0.001, 0.01, 0.1, 1 and 10
        \subref{fig : 15b} RNN model with learning rates $\alpha$ at 0.001, 0.01, 0.1, 1 and 10}
\label{fig : CNN-RNN }
\end{figure}
In Fig. \ref{fig : CNN-RNN } we plot after-state values for the CNN architecture (Fig.\ref{fig:Network 2}) and the RNN architecture (Fig.\ref{fig:Network 3}) with batch training. The CNN model takes 50 randomly sampled states that it has encountered while the RNN model uses sequences of 10 states. The Q-learning algorithm is used for target updates and RMSPROP algorithm is used to optimize with the network parameters. We show results for 5 learning rate parameters $\alpha$.\par
\begin{figure}
\centering
\subfigure[]{\label{fig : 16a}\includegraphics[scale=0.15]{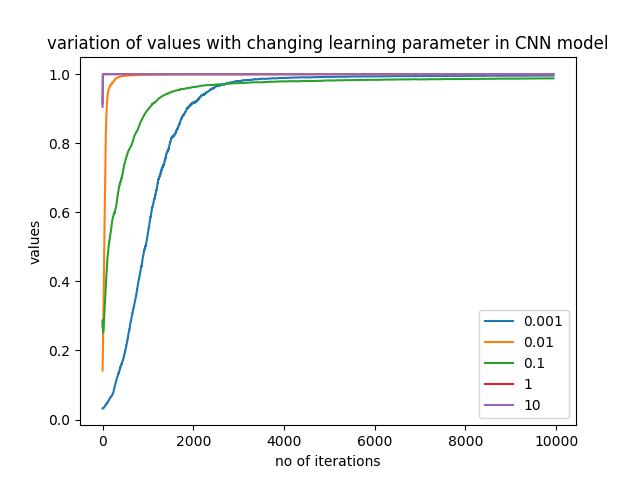}}\hspace{2pt}
\subfigure[]{\label{fig : 16b}\includegraphics[scale=0.15]{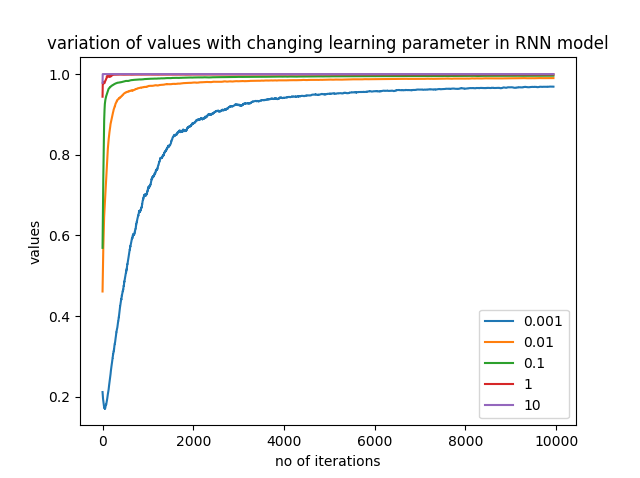}}
\caption[comparison of CNN and RNN model]{After state values implemented using SARSA and batch training for 
        \subref{fig : 16a} CNN model with learning rates $\alpha$ at 0.001, 0.01, 0.1, 1 and 10
        \subref{fig : 16b} RNN model with learning rates $\alpha$ at 0.001, 0.01, 0.1, 1 and 10}
\label{fig : CNN-RNN_SARSA }
\end{figure}
In Fig. \ref{fig : CNN-RNN_SARSA } we use the SARSA algorithm instead of Q-learning to compare the after-state values using the two different architectures of the CNN (Fig. \ref{fig:Network 2}) and RNN (Fig. \ref{fig:Network 3}) using batch training similar to the one used to generate Fig. \ref{fig : CNN-RNN } for five learning rate parameters $\alpha$.\par
\begin{figure}
\centering
\subfigure[]{\label{fig : 17a}\includegraphics[scale=0.15]{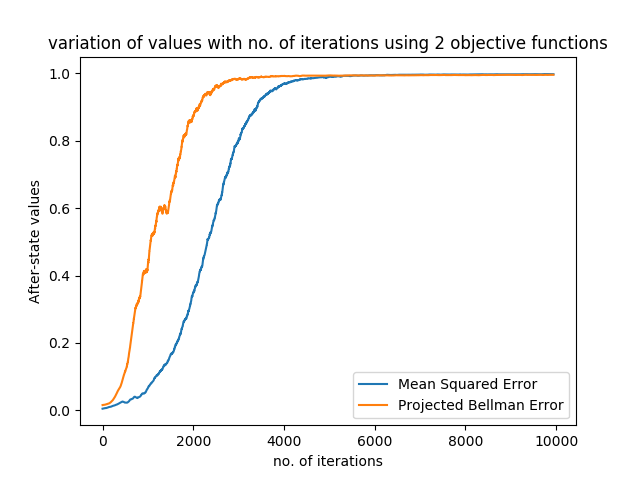}}\hspace{2pt}
\subfigure[]{\label{fig : 17b}\includegraphics[scale=0.15]{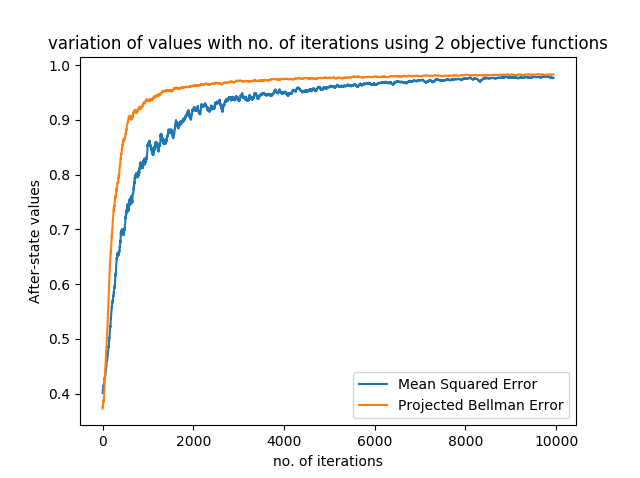}}
\caption[comparison of PBE-VE-values]{After-state values using 2 different objective functions for 
        \subref{fig : 17a} CNN architecture
        \subref{fig : 17b} RNN architecture}
\label{fig : PBE-VE-values}
\end{figure}

\begin{figure}
\centering
\subfigure[]{\label{fig : 18a}\includegraphics[scale=0.15]{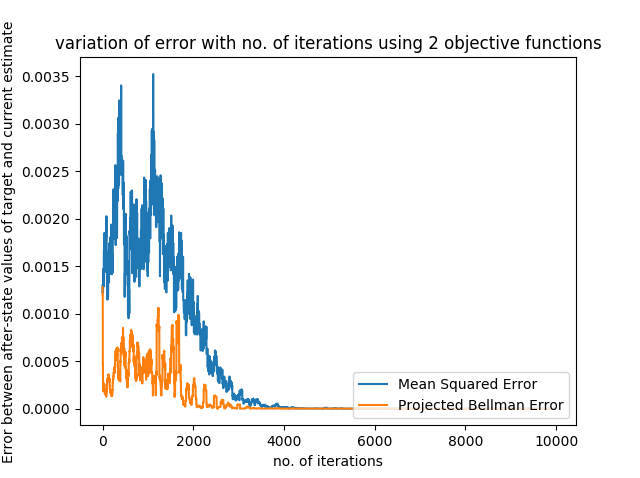}}\hspace{2pt}
\subfigure[]{\label{fig : 18b}\includegraphics[scale=0.15]{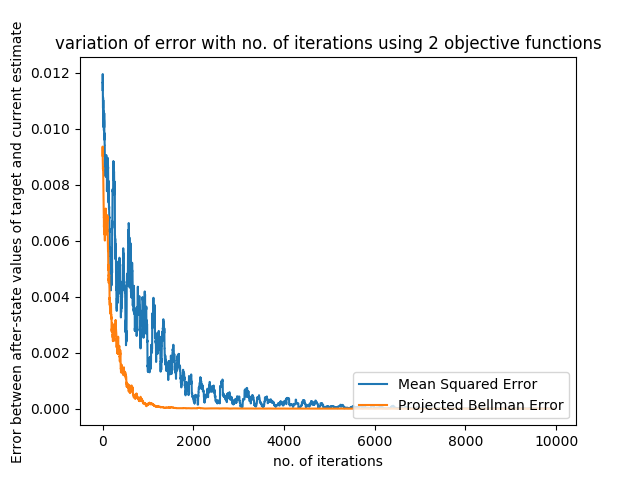}}
\caption[comparison of PBE-VE-costs]{Training Errors using 2 different objective functions for 
        \subref{fig : 18a} CNN architecture
        \subref{fig : 18b} RNN architecture}
\label{fig : PBE-VE-costs }
\end{figure}

In Fig. \ref{fig : PBE-VE-values} we plot the After-State values learned using Mean Squared Error and Projected Bellman Error (as we have implemented) as objective functions and targets estimated by Q-learning and SARSA respectively for both the CNN and RNN architectures. For the MSE error we have kept the learning rate at 0.001 and for the PBE error we have used two different learning rates. For the projection operation we have kept the learning rate at 0.1 and for the parameter update we have kept the learning rate at 0.001. To ensure that network parameters are kept independent for both the updates we have used 2 identical networks, one for performing the projection operation and the other for value function estimation. Both networks are updated based on the two error functions as described. In Fig. \ref{fig : PBE-VE-costs } we plot the network-training errors using Mean Squared Error and Projected Bellman Error (as we have implemented) as objective functions for both CNN and RNN architecture. \par
\begin{figure}
    \centering
    \includegraphics[scale = 0.15]{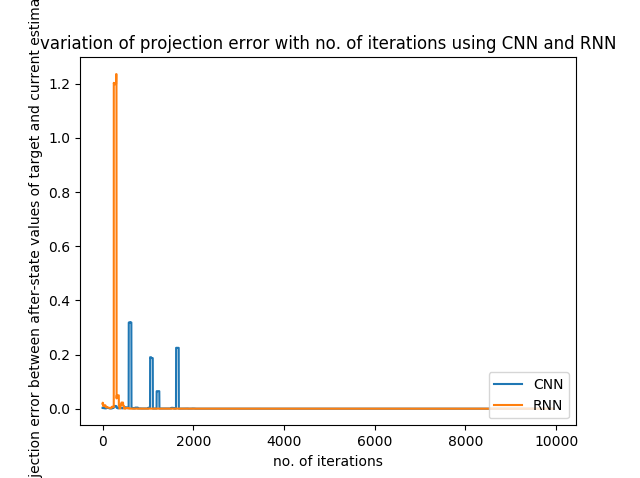}
    \caption{Comparisons of the Projection Error using RNN and CNN}
    \label{fig:Projection_Error}
\end{figure}

  
In Fig. \ref{fig:Projection_Error} we plot the projection error that is incurred between the target and the best representable function for both architectures.\par
\begin{figure}
    \centering
    \includegraphics[scale = 0.26]{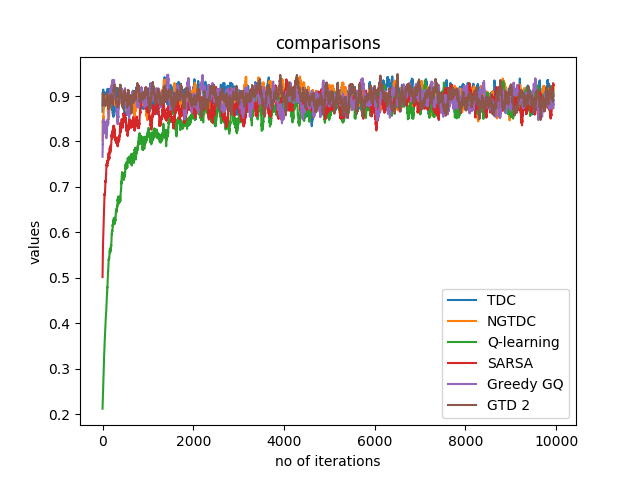}
    \caption{Comparisons of the different algorithms}
    \label{fig:Comparisons}
\end{figure}

In Fig. \ref{fig:Comparisons} we compare between all 6 algorithms using a network depicted in Fig. \ref{fig:Network 1} using online updates and using best hyperparameters found in our work.\par

To determine how well these algorithms perform in actual game playing, we created four agents. CNN-MSE is a 4-layer convolution architecture and uses the Mean Square Error as the objective function that the network tries to minimize and Q-learning to update the after-state value targets. CNN-PBE is a 4-layer convolution architecture and uses the Projected Bellman Error as the objective function that the network tries to minimize and uses SARSA to update the after-state value targets. RNN-MSE is a 3-layer convolution followed by a RNN layer architecture and uses the Mean Square Error as the objective function that the network tries to minimize and Q-learning to update the after-state value targets. RNN-PBE is a 3-layer convolution followed by a RNN layer architecture and uses the Projected Bellman Error as the objective function that the network tries to minimize and uses SARSA to update the after-state value targets. We play for 2 seasons, with each season holding 10 tournaments and each tournament having a total of 12 games with the 4 players playing against each other. In each tournament a single player plays 3 games as the first player and 3 games as the second player. In each season a single player plays 30 games as the first player and 30 games as the second player. The results are shown in Table \ref{table : table1}.

\begin{table*}[]
\centering
\caption{Final Resuls}
\label{table : table1}
\resizebox{\textwidth}{!}{%
\begin{tabular}{|c|l|c|c|c|c|c|c|c|c|}
\hline
\multicolumn{2}{|c|}{Season}     & \multicolumn{2}{c|}{CNN-MSE} & \multicolumn{2}{c|}{CNN-PBE} & \multicolumn{2}{c|}{RNN-MSE} & \multicolumn{2}{c|}{RNN-PBE} \\ \hline
\multicolumn{2}{|c|}{}           & First Player & Second Player & First Player & Second Player & First Player & Second Player & First Player & Second Player \\ \hline
\multicolumn{2}{|c|}{1}          & 20           & 7             & 24           & 14            & 20           & 8             & 19           & 8             \\ \hline
\multicolumn{2}{|c|}{2}          & 23           & 8             & 23           & 12            & 21           & 5             & 22           & 7             \\ \hline
\multicolumn{2}{|c|}{TOTAL WINS} & \multicolumn{2}{c|}{58}      & \multicolumn{2}{c|}{73}      & \multicolumn{2}{c|}{54}      & \multicolumn{2}{c|}{56}      \\ \hline
\end{tabular}%
}
\end{table*}

\section{Conclusion}
By glancing through the results in Table \ref{table : table1} one immediate conclusion that can be drawn is using an extra Convolution Layer instead of a RNN layer improves overall performance thus leading us to believe our original intuition about better improvements was wrong. This is however not surprising as Hex is a complete information game. At any instance of a board one has a winning strategy which is independent of how the game was played before \cite{wikiHex}. RNN would perhaps be better suited in games where one needs to remember the moves played in sequence like many card games. The next obvious result is that our implemented error function which we call PBE performs better than the MSE in both models. Our hypothesis is that the HEX game can be better modeled with a CNN architecture than using an RNN architecture as can be seen from the projection error graph in Fig. \ref{fig:Projection_Error}. Further we would like to hypothesize that with improved function approximators (in our case CNN), the Projected Bellman Error performs significantly better than when applied on poor function approximators. Also that our error function is very similar to the GTD2 and TDC algorithms as derived by Sutton et al, can be noticed in how we also use two learning rates and the faster convergence exhibited by both figures where one is updated manually using the GTD2 and TDC updates and the other through an explicit given error function that we have proposed (see for example Figs. \ref{fig:Comparisons} and \ref{fig : PBE-VE-values}). Though we have not formally proved the similarity, we leave it for future work. 

\section{Future Work}
A formal investigation into calculating the actual gradients of the objective function we have proposed and studying the convergence guarantees is still to be done. Also as suggested in \cite{seijen2014true} the idea of eligibility traces as applied has to be further investigated. Finally the performance in a large $13 \times 13$ board against strong human players has to be conducted before making further comments.

\section{Acknowledgement}
Heartfelt gratitude towards Prof. Shalabh Bhatnagar, Prasenjit Karmakar, Chandramouli Kamanchi, Abhik Singla and the Stochastic Systems Lab. team. Special token of respect towards Prof Sastry and the EECS faculty of Indian Institute of Science. Shib Shankar Dasgupta, Kisnsuk Sircar, Swayambhu Nath Ray have been indispensable during the hard times. A very general gratitude towards Indian Institute of Science and the great nation of India. A special note towards Priyaam Roy for her help and support. A final note of thank you towards my parents whose constant well wishes were with me always.

%



\ifCLASSOPTIONcaptionsoff
  \newpage
\fi

\bibliographystyle{IEEEtran}

\bibliography{mybib}

%


\end{document}